\documentclass[sigconf, nonacm]{acmart}

\usepackage{algorithm}
\usepackage{algpseudocode}
\usepackage{listings}
\usepackage{float}
\usepackage{comment}
\usepackage{tabularx}
\usepackage{booktabs}
\usepackage{stfloats}
\usepackage{multirow}
\usepackage{placeins}

\AtBeginDocument{%
  }

\copyrightyear{2026}
\acmYear{2026}
\setcopyright{rightsretained}
\acmConference[FDG '26]{International Conference on the Foundations of Digital Games}{August 10--13, 2026}{Copenhagen, Denmark}
\acmBooktitle{International Conference on the Foundations of Digital Games (FDG '26), August 10--13, 2026, Copenhagen, Denmark}
\acmPrice{}
\acmDOI{XXXXXXX.XXXXXXX}
\acmISBN{/26/08}

\begin{document}

\title{High Dimensional Procedural Content Generation}

\author{Kaijie Xu}
\affiliation{%
  \institution{McGill University}
  \city{Montreal}
  \state{Quebec}
  \country{Canada}}
\email{kaijie.xu2@mail.mcgill.ca}

\author{Clark Verbrugge}
\affiliation{%
  \institution{McGill University}
  \city{Montreal}
  \state{Quebec}
  \country{Canada}}
\email{clump@cs.mcgill.ca}

\begin{abstract}
Procedural content generation (PCG) has made substantial progress in shaping static 2D/3D geometry, while most methods treat gameplay mechanics as auxiliary and optimize only over space. We argue that this limits controllability and expressivity, and formally introduce High-Dimensional PCG (HDPCG): a framework that elevates non-geometric gameplay dimensions to first-class coordinates of a joint state space. We instantiate HDPCG along two concrete directions. Direction-Space augments geometry with a discrete layer dimension and validates reachability in 4D $(x,y,z,\ell)$, enabling unified treatment of 2.5D/3.5D mechanics such as gravity inversion and parallel-world switching. Direction-Time augments geometry with temporal dynamics via time-expanded graphs, capturing action semantics and conflict rules. For each direction, we present three general, practicable algorithms with a shared pipeline of abstract skeleton generation, controlled grounding, high-dimensional validation, and multi-metric evaluation. Large-scale experiments across diverse settings validate the integrity of our problem formulation and the effectiveness of our methods on playability, structure, style, robustness, and efficiency. Beyond quantitative results, Unity-based case studies recreate playable scenarios that accord with our metrics. We hope HDPCG encourages a shift in PCG toward general representations and the generation of gameplay-relevant dimensions beyond geometry, paving the way for controllable, verifiable, and extensible level generation.
\end{abstract}

\ccsdesc[500]{Applied computing~Computer games}

\keywords{Procedural Content Generation, Level Generation, Time Expanded Graph, A* Search, Dynamic Programming, Genetic Algorithms}

\maketitle

\section{Introduction}

Procedural content generation (PCG) automates the production of game assets and levels so designers can focus on higher-level intent while still achieving variety, replayability, and scale \cite{shaker2016procedural, togelius2011search, summerville2018procedural}. Despite decades of progress, the dominant formulations for level generation remain largely geometry-first: 2D tile maps and aesthetically oriented 3D scenes are common \cite{jiang2022learning, earle2024dreamcraft, hu20243d, earle2025dreamgarden, kobenova2024social, xu2025constraint, xu2025database}, while mechanics such as time-dependent traversal, discrete interaction rules, and additional non-spatial state are injected via simulation-based objective functions or post-processing heuristics rather than being represented natively in the generator \cite{smith2010tanagra, yannakakis2011experience, shaker2013evolving, stephenson2016procedural, ferreira2014generating, fernandes2024generating}. While simulation is essential for continuous domains, it typically enforces such requirements only indirectly, making it harder to target specific mechanism-level structure during generation. This gap suggests a missing general, extensible representation and problem formulation for complex, mechanism-rich PCG.

To address this gap, we study High-Dimensional PCG (HDPCG) tasks and propose a general representation and pipeline that augments spatial nodes with additional gameplay dimensions (e.g., layers \cite{van2011navigation, hans2020spaces}, time \cite{witkin1988spacetime, zagal2010time}, locomotion modes and style \cite{al2018virtual, anderton2025teleportation, kovar2023motion, liu2005learning}). HDPCG encodes the world state as attribute-labeled cells over an expanded state space, enabling planners to reason about geometry and dynamics in a single expanded graph. Conceptually, HDPCG combines classic geometric PCG with structured representations for game rules and temporally evolving elements, drawing on ideas from dynamic networks/time-expanded graphs to handle evolution over time via graph expansion when appropriate \cite{hoppe2000quickest, kohler2002time, wang2019time, erdmann1987multiple, xu2025actionwindow}. Recent systems that treat level design as metric-driven optimization \cite{cooper2023sturgeon, cooper2024sturgeon, cooper2025constraint} motivate this direction; HDPCG generalizes these ideas into a unified high-dimensional state-space formulation.

To validate breadth and practicality, we test HDPCG along two representative axes that mirror modern game design: Direction-S (Space), which augments geometry with discrete layers and controlled pacing; and Direction-T (Time), which augments geometry with temporal dynamics and proves feasibility on a time-expanded graph. For each direction, we clearly define the problem, provide multiple algorithmic families, and evaluate them under common metrics for controllability, robustness, efficiency, and quality. Finally, we validate playability and faithfulness via Unity case studies, replaying our computed solutions in-engine as layered gravity-flip levels similar to the single-button gravity inversion in \emph{VVVVVV}\footnote{\url{https://thelettervsixtim.es}}, and as parallel-timeline puzzles where the player switches between two time states to traverse geometry and enemies, in the spirit of \emph{Dishonored 2: A Crack in the Slab}\footnote{\url{https://bethesda.net/en/game/dishonored2}} and \emph{Titanfall 2: Effect and Cause}\footnote{\url{https://www.ea.com/games/titanfall/titanfall-2}}. For the Time direction we additionally realize moving-platform sequences with tight timing windows, similar to 3D platform-jumping challenges in games like \emph{Super Mario 3D World}\footnote{\url{https://supermario3dworld.nintendo.com/}}. Together, these results position HDPCG as a basis for future PCG that moves beyond geometry-only synthesis towards mechanism- and dynamics-aware level generation, expanding the design space while maintaining controllability and analytical rigor. Our contributions are:
\begin{itemize}
    \item We propose HDPCG, a general, extensible representation and problem formulation for high-dimensional level generation, integrating geometry, time, and other gameplay axes under a single graph framework.
    \item We instantiate HDPCG in two concrete directions, Space and Time, each with a group of generic methods.
    \item We conduct multi-scale experiments with unified metrics that quantify controllability, robustness, efficiency, and composite quality, enabling fair cross-method comparisons.
    \item We deliver Unity case studies for both directions, showing faithful mechanism reproduction and end-to-end feasibility from HDPCG plans to real-time 3D gameplay.
\end{itemize}
\section{Related Works}


Early level PCG has largely been geometry-first, representing levels as spatial grids while treating mechanics and dynamics (e.g., timing and ruleful interactions) as secondary, often handled via fitness/discriminators, post-processing, or simulation when needed (especially for physics-rich games) \cite{togelius2011search, summerville2018procedural, parish2001procedural, wonka2003instant}. As a result, these pipelines often struggle when layout and core mechanics are tightly interdependent \cite{yannakakis2011experience, dormans2011generating, dormans2011level, fernandes2024generating}.
More recent work pushes beyond static tiles toward PCG that must respect complex dynamics, richer 3D structure, or temporal constraints. 
Mixed-initiative systems such as Tanagra link layout to pacing and jump physics via reactive planning/constraint solving, but remain on 2D grids with mechanics encoded as constraints over geometry rather than as state variables \cite{smith2010tanagra}, which makes the mechanic state less reusable across different dynamics and harder to certify via a single unified planner.
Rhythm-based generators similarly derive geometry from action rhythms, yet time is an external driver rather than a dimension of the generative state \cite{smith2009rhythm, smith2010launchpad}. 
For 3D levels, controllable generators learn to place volumetric chunks and structural affordances, but typically leave detailed mechanic interactions to downstream checks or reward shaping \cite{jiang2022learning}. 
Beyond platformers, physics-based puzzle generation couples geometry with continuous dynamics and timing, e.g., Cut the Rope and Angry Birds families ensure solvability under simulation \cite{shaker2013evolving, van2015procedural, stephenson2016procedural, ferreira2014generating}; stealth generation integrates moving guards/cameras with patrol timing and visibility fields \cite{xu2014generative}. 
LLM-assisted 3D pipelines now target functional, controllable worlds: DreamCraft synthesizes text-guided Minecraft environments with explicit constraints \cite{earle2024dreamcraft}, T2BM uses LLM prompting for multi-storey buildings with functional blocks \cite{hu20243d}, and DreamGarden supports designer-in-the-loop growth of Unreal scenes from a single prompt \cite{earle2025dreamgarden}. Social Conjuring explores multi-user, runtime co-creation of dynamic 3D scenes \cite{kobenova2024social}. Closer to full 3D level pipelines, optimization-based approaches use LLMs plus constraints to assemble multi-floor structures and then furnish mechanics \cite{xu2025constraint, xu2025database}. Finally, a line of work grounded in WaveFunctionCollapse \cite{karth2017wavefunctioncollapse, karth2021wavefunctioncollapse} explicitly explores time as a modeling axis \cite{zagal2010time}: Facey and Cooper operate on $(x,y,t)$ blocks to generate levels and playthroughs \cite{facey2024toward}, and a follow-up generalizes this spacetime level generation \cite{Vandara2025SpacetimeLevelGenerationEditing}. Complementary to these WFC-style models, the Sturgeon family of systems treats level and playthrough design as constraint-based multi-objective optimisation: Sturgeon-MKIII and MKIV jointly enforce tile and graph label rewrite rules together with designer-specified metrics over example paths, while a related formulation frames level creation as multi-objective optimization over playthrough statistics \cite{cooper2023sturgeon, cooper2024sturgeon, cooper2025constraint}. Together, these examples show rising interest in integrating geometry with dynamics and temporal logic, while many methods still treat these aspects as add-ons to basic geometric representations.

Our approach moves forward by lifting mechanics into the representation itself. Concretely, we model generation on a dimensional expanded graph (DEG) in which spatial nodes are augmented with additional discrete axes (e.g., time steps, world ``phases,'' puzzle states, etc.). This idea parallels time-expanded networks from dynamic flow theory, where a static graph is ``unrolled'' across time to reason about feasibility, providing well-studied guarantees and algorithmic tools \cite{hoppe2000quickest, kohler2002time, wang2019time, witkin1988spacetime}. It is also conceptually aligned with configuration-space-time in motion planning, where obstacles and capabilities are embedded directly into a higher-dimensional search space to ensure realizability of trajectories \cite{erdmann1987multiple}. Related mission planning works similarly cast patrol schedules as action windows that constrain feasible paths, reinforcing the need to make time an explicit axis of the state space \cite{xu2025actionwindow}. In our framework, mechanical feasibility is checked during generation via graph search, allowing geometry and mechanics to co-evolve and making solvability a first-class invariant. This contrasts with constraint-driven approaches that optimize over geometric encodings and only indirectly capture mechanical interdependence. Moreover, we detail how DEG layers are constructed and validated via search over the expanded graph, operationalizing this theory for high-dimensional PCG.

\section{Method}
\label{sec:method}

This section formalizes HDPCG on a DEG, introduces a unified four stage pipeline, and details its instantiation in two directions via concrete planning, validation, and search methods. In Space instantiation, the player explores a level made of two overlapping layers and can only switch under rules; in Time instantiation, the player must time movement by waiting and crossing while platforms and enemies move on a repeating cycle.

\subsection{General method Definition and Pipeline}
\label{sec:method:general}

\paragraph{High‐dimensional level representation:}
We model a level as a finite discrete spatial grid $\mathcal{X}\subset\mathbb{Z}^d$ ($d\!\in\!\{2,3\}$) together with $k$ gameplay dimensions
$\mathcal{D}^1,\ldots,\mathcal{D}^k$ (e.g., parallel layers, time, locomotion modes).
The overall state space is
\[
\mathcal{S}\;=\;\mathcal{X}\times\mathcal{D},\qquad
\mathcal{D}\;=\;\mathcal{D}^1\times\cdots\times\mathcal{D}^k.
\]
A \emph{cell} is $s=(x,d)\in\mathcal{S}$ with $x\in\mathcal{X}$ and $d\in\mathcal{D}$.

\paragraph{Attribute-valued occupancy:}
\label{def:apf}
We use an attribute labeling
$
A:\mathcal{S}\!\to\!\mathcal{A},
$
where $\mathcal{A}$ contains both static and dynamic gameplay symbols (e.g., empty, wall, enemy, platform, and their dimension-aware variants such as moving\_enemy@t).
A cell's traversability is given by a predicate
$
P(s;\,A,\upsilon)\in\{0,1\}
$
that may depend on agent capability $\upsilon$ and the label $A(s)$.
Likewise, transition feasibility is controlled by
$
F(s,s';\,A)\in\{0,1\}
$
to express constraints such as collision, swap prevention, or layer-switch rules.

\paragraph{Adjacency and costs:}
\label{def:edgecost}
Let $\mathcal{A}_{X}\subset\mathbb{Z}^d\!\setminus\!\{0\}$ be the spatial neighborhood (4/6-connected) and, for each gameplay dimension $j$, let $\mathcal{A}_j\subseteq \mathcal{D}^j\!\times\!\mathcal{D}^j$ be admissible local transitions (e.g., layer switch $(\ell,\ell')$, time successor $(t,t\!+\!1)$).
We define the directed edge set
\[
\mathcal{E}\;=\;\big\{(s,s'):\; 
\big(x'-x\in\mathcal{A}_{X}\wedge d'=d\big)\ \lor\
\big(x'=x\wedge \exists j:\ (d^j,d'^j)\!\in\!\mathcal{A}_j\big)\big\}
\]
and retain edges with $P(s)\!=\!P(s')\!=\!1$ and $F(s,s')\!=\!1$. 
For axes that must advance each step (e.g.\ time), we disallow same-step spatial moves and couple moves with the dimension transition (e.g.\ $(x,t)\to(x',t+1)$ or $(x,t)\to(x,t+1)$).
A nonnegative edge cost $c:\mathcal{E}\!\to\!\mathbb{R}_{\ge 0}$ assigns unit cost to spatial moves and dimension-specific costs to non-spatial transitions (e.g., layer-switch penalty $\lambda$), and may be modulated by attributes via $c(e;A)$.

\paragraph{Dimensional-Expanded Graph:}
\label{def:deg}
For ordered dimensions (e.g., time) or to bound search, we instantiate a finite subgraph
\[
\widetilde{\mathcal{G}}=\big(\widetilde{\mathcal{S}},\widetilde{\mathcal{E}}\big),\quad
\widetilde{\mathcal{S}}\subseteq\mathcal{S},
\]
by truncating along those axes (e.g., $t\!\le\!T_{\max}$) and inducing edges from $\mathcal{E}$.
The classical time-expanded graph (TEG) is the special case with $\mathcal{D}^1\!=\!\{0,\ldots,T_{\max}\}$ and $\mathcal{A}_1\!=\!\{(t,t\!+\!1)\}$.
Our Space direction expands along discrete \emph{layers}; our Time direction expands along \emph{time}.
Dynamic elements such as moving enemies or platforms are represented directly by $A(s)$ varying across the expanded dimension and by the feasibility predicate $F$.

\paragraph{High-dimensional problem definition:}
\label{def:problem}
Given start $s_{\mathrm{start}}\!\in\!\widetilde{\mathcal{S}}$, a goal set $\mathcal{G}_{\mathrm{goal}}\!\subseteq\!\widetilde{\mathcal{S}}$ (typically all states with spatial coordinate $x_g$), and generator parameters $\theta$ producing $(A,c)$ (and optional priors/anchors), synthesize a level instance and a feasible path
\[
\pi^\star \in \arg\min_{\pi\in\Pi(A,c)}\ C(\pi)\quad
\text{s.t.}\ \pi: s_{\mathrm{start}}\leadsto \mathcal{G}_{\mathrm{goal}},\ \mathbf{m}(\pi,A)\in\mathcal{K},
\]
where $C(\pi)=\sum_{e\in\pi} c(e;A)$, $\Pi(A,c)$ is the set of admissible paths in $\widetilde{\mathcal{G}}$, $\mathbf{m}$ collects structural/playability metrics (coverage, tortuosity, switch-pattern compliance, robustness, etc.), and $\mathcal{K}$ encodes target ranges.
A scalarized score $J(\pi,A)=w^\top \mathbf{m}(\pi,A)$ is used when convenient. Validation, robustness probing, and metric computation are applied uniformly on the DEG across all instantiated directions.

\paragraph{General solution scheme:}
We adopt a general four-stage scheme that unifies Space/Time (and other) dimensions:
\begin{enumerate}
  \item \emph{Abstract planning in $\mathcal{X}$ or directly in $\widetilde{\mathcal{G}}$:}
  compute a sparse route using a shaped spatial potential
  $
  c_{\mathrm{sp}}(x)=c_0(x)+\alpha\,\phi(x)-\beta\,\psi(x),
  $
  where $\phi$ encodes repulsion (e.g., obstacle blobs, dispersion around laid segments) and $\psi$ encodes attraction to targeted features (e.g., intended layer-switch coordinates or time-synchronized events) \cite{khatib1986real}.
  \item \emph{Instantiation of attributes and rules:}
  expand the abstract plan into attribute-valued occupancy and transition rules, yielding $A$ and $F$ (e.g., corridor/room carving across layers; time-varying labels for moving enemies/platforms; swap/collision constraints), and assemble the edge costs $c$ on $\widetilde{\mathcal{G}}$.
  \item \emph{Validation by shortest-path search:}
  run A* \cite{hart1968formal}, Breadth First Search (BFS), or Dynamic Programming (DP) on $\widetilde{\mathcal{G}}$ to obtain a feasible path $\pi$; optionally replan under localized perturbations for robustness. If validation fails, we treat the instance as invalid, reject it in Space, or resample the generator parameters and retry in Time (e.g., object placement/catalog).
  \item \emph{Evaluation and (optional) search over $\theta$:}
  compute $m(\pi,A)$ and either accept the instance or update $\theta$ (e.g., evolutionary operators on anchors \cite{holland1992adaptation}, potentials, or object sets) to improve $J(\pi,A)$ subject to $\mathcal{K}$.
\end{enumerate}
This dimensional-expanded view treats geometry and auxiliary gameplay axes within a single graph-theoretic framework, enabling principled comparison and optimization across heterogeneous high-dimensional PCG tasks.

\subsection{Direction-S: Space}
\label{sec:method:space}

\subsubsection{Problem Setup and Modeling}
We inherit all notation and definitions from Section~\ref{sec:method:general}. In the Space direction we specialize the gameplay dimension to a finite layer set $\mathcal{L}$ and take the DEG over $\mathcal{X}\times\mathcal{L}$. Spatial moves and layer transitions are exactly those in Section~\ref{def:edgecost} with a nonnegative switch penalty $\lambda$. The only Space-specific additions are two feasibility policies encoded via $A$ and $F$: (i) cross-layer mutual exclusion at the same spatial coordinate (non-active layers are blocked unless the coordinate is a designated switch site), and (ii) switch consistency that opens the landing layer and locally seals the departure side to prevent immediate back-leak.

\subsubsection{Abstract Skeleton and Controlled Geometry}
Stage~1 (Skeleton Planning): we compute a route in $\mathcal{X}$ under the spatial potential
\[
c_{\mathrm{sp}}(x)=c_0(x)+\alpha\,\phi(x)-\beta\,\psi(x),
\]
where $\phi$ promotes dispersion (e.g., repulsion around carved segments or repulsive blobs) and $\psi$ attracts the route toward intended switch features (when present) \cite{khatib1986real, rimon1992exact}. The result is a polyline $\langle x_0,\ldots,x_M\rangle$ which we annotate with layer labels to form a skeleton $\Sigma=\langle(x_i,\ell_i)\rangle_{i=0}^M$ (indices or coordinates may mark switches).

Stage~2 (Controlled Grounding): we instantiate attributes and rules by two-pass \emph{corridor$\to$room} growth around $\Sigma$ with component-aware conflict checks. At each designated switch $(x,\ell\!\to\!\ell')$ we set $A(x,\ell')=\texttt{empty}$, and enforce $F$ to (a) allow the same-coordinate layer edge $(x,\ell)\!\to\!(x,\ell')$ and (b) discourage immediate reversal by sealing the departure side in a small neighborhood.

Stage~3 (High-Dimensional Validation): we validate reachability on the DEG specialized to layers (Section~\ref{def:deg}) and run A* on that graph with the Manhattan heuristic
$
h(x,\ell)=\|x-x_g\|_1,
$
which is admissible/consistent since switches are nonnegative-cost and do not decrease spatial distance. Search terminates on $\{(x_g,\ell):\ell\in\mathcal{L}\}$; the recovered path $\pi$ feeds evaluation and (optional) robustness replanning.

\subsubsection{Methods}
We consider three increasingly controllable instantiations that differ only in how $c_{\mathrm{sp}}$ and switch sites are chosen; Stage~2/3 remain identical.

\paragraph{\textbf{Method 1: Naive Noise Baseline}}
A minimally guided reference with index-based switching. We take $c_0\equiv 1$ on open cells and draw an i.i.d.\ noise field $\eta:\mathcal{X}\to[0,1]$; set
\[
c_{\mathrm{sp}}(x)=1+\alpha\,\eta(x)\qquad(\phi\equiv 0,\ \psi\equiv 0),
\]
run a single 3D A* to obtain $\langle x_0,\ldots,x_M\rangle$, and choose $K$ interior indices at near-uniform spacing (with slight jitter) to flip layers: $\ell_i=1-\ell_{i-1}$ while keeping $x_i=x_{i-1}$. Stage~2 realizes geometry and rules; Stage~3 validates with 4D A*. In Genetic Algorithms (GA) mode, an individual encodes the noise seed, the switch count $K$ and its index set along the backbone, plus corridor/room widths; mutation jitters indices, flips $K\!\pm\!1$, and perturbs widths, and fitness is the scalarized $J(\pi,A)$ after validation.

\paragraph{\textbf{Method 2: Naive Penalty A*}}
Dispersion-aware planning with still index-based switching. We plan segment-wise between waypoints $\{w_r\}$ using
\[
c_{\mathrm{sp}}(x)=c_0(x)+\alpha\,\phi(x),\qquad \psi\equiv 0,
\]
and accumulate $\phi$ after each segment via a short-range kernel around newly carved voxels, gently steering later segments away from earlier ones. After concatenation, we sample $K$ non-waypoint indices with a loose minimum-gap constraint and flip layers as above. Geometry instantiation and 4D validation follow unchanged. In GA mode, an individual encodes the dispersion field parameters (kernel radius/decay and the repulsion weight $\alpha$), the waypoint seeds, and the index sampler for $K$ switches; mutation jitters waypoints, perturbs kernel/weight, and resamples a small subset of switch indices.

\paragraph{\textbf{Method 3: Potential Field A*}}
Explicit spatial targeting of switch \emph{coordinates}. We prescribe a set $\mathcal{Q}=\{q_1,\ldots,q_K\}\subset\mathcal{X}$ (min-spacing enforced) and define
\[
c_{\mathrm{sp}}(x)=c_0(x)+\alpha\,\phi_{\text{blob}}(x)-\beta\,\psi_{\text{switch}}(x),
\]
where $\phi_{\text{blob}}$ aggregates repulsive blobs (discouraging undesirable regions) and $\psi_{\text{switch}}$ provides narrow attraction at points in $\mathcal{Q}$. The resulting 3D A* polyline tends to thread $\mathcal{Q}$; we flip layers precisely at the indices where $x_i\in\mathcal{Q}$ (excluding endpoints). If any stochastic operator yields a cross-layer “teleport” (layer change across different coordinates), we locally patch by re-planning within the current layer to the intended coordinate before switching. Stage~2 enforces switch consistency in $A,F$; Stage~3 completes validation. In GA mode, an individual encodes switch anchors (waypoints) and potential-field parameters (blob locations/radii and weights); mutation applies coordinate jitter and radius/weight noise.

\subsection{Direction-T: Time}
\label{sec:method:time}

\subsubsection{Problem Setup and Modeling}
We inherit all notation from Section~\ref{sec:method:general}. The Time direction specializes the DEG to either a finite-horizon or cyclic time axis. We instantiate feasibility $F$ to allow three transition types under time-varying attributes $A$: (i) wait $(x,t)\to(x,t+1)$ when no violation occurs, (ii) walk $(x,t)\to(x',t+1)$ for $x'-x\in\mathcal{A}_X$ if the landing cell is open at $t+1$, and (iii) ride along a platform's motion with endpoint-only boarding/alighting and intermediate occupancy checks; we also prevent head-on swaps with moving obstacles \cite{van2011reciprocal}. Edge costs follow Section~\ref{def:edgecost} (unit spatial steps with optional weights for wait/ride).

\subsubsection{Validation on TEG}

Validation follows the general shortest-path scheme on the DEG, using A* when costs are non-uniform and BFS/DP otherwise. We employ the admissible heuristic
$
h(x,t)=\|x-x_g\|_1,
$
and choose either a finite horizon $T_{\max}$ (truncation) or a cyclic domain $\mathbb{Z}_T$ (time wraps modulo $T$). 

\subsubsection{Methods}
We provide three increasingly expressive instantiations. All share Stage~2/3 (attribute instantiation $A$ and rule set $F$, followed by TEG validation); they differ in Stage~1 (how we plan in $\mathcal{X}$ vs.\ directly in $\widetilde{\mathcal{G}}$ and how dynamic elements are proposed).

\paragraph{\textbf{Method 1: Static Backbone Baseline}}
Stage~1 computes a spatial backbone in $\mathcal{X}$ using 2D/3D A* under $c_{\mathrm{sp}}(x)=c_0(x)$ (optionally a lightly shaped variant). Around this backbone we instantiate $A$ by placing periodic platforms and obstacles (periods, phases, ranges drawn from priors), and derive $F$ to enforce head-on swap prevention and endpoint-only boarding. Feasibility is then validated by BFS on the induced TEG. If infeasible, object parameters are resampled and validation is retried. This baseline captures a \emph{plan-then-animate} workflow with minimal time awareness in Stage~1. In GA mode, an individual encodes the dynamic object catalog (platform counts, spans, periods and phases; obstacle patrols) optionally tied to the spatial-backbone seed; mutation resamples a subset of object parameters and applies phase shifts.

\paragraph{\textbf{Method 2: TEG-A*}}
An ideal validator would run an exact A* on the full time-expanded graph $\widetilde{\mathcal{G}}$, with $(x,t)$ states and transitions for \emph{wait}/\emph{walk}/\emph{ride}, endpoint-only boarding/alighting, intermediate occupancy checks, and swap-conflict prevention. To actively drive meaningful interactions rather than incidental ones, the exact variant (as a witness-search objective for certification/metrics) must also reward riding edges 
and store interaction state (e.g., which platforms/endpoints have been used) so that repeated, low-value contacts are not mistaken as progress. This, however, leads to a combinatorial blow-up: the state must carry interaction masks or counters on top of $(x,t)$, and the branching factor grows with dynamic elements. In practice we observed timeouts even on our smallest grids under such a formulation.

For tractability, we adopt a simplified TEG-A* baseline. We operate on a cyclic TEG ($\mathbb{Z}_T$) and keep feasibility rules intact: endpoint-only riding, per-step occupancy and swap checks, and disallowing mid-track walking-while using a coarse cost model (unit costs for walk/wait; ride cost equals ride duration) and lightweight bookkeeping (bitmasks for “used platforms” and “visited obstacle endpoints”). We employ $h(x,t)=\|x-x_g\|_1$ and a single-close policy (no re-opening) to keep search compact. This pragmatic baseline validates time-consistent paths efficiently but, being a shortest-path search without strong temporal shaping, does not reliably induce rich interactions even when constraints are present. We nevertheless include it for completeness; the observed limitations motivate our choice of \textbf{TEG-DP} as the main solver, which better balances scalability and temporal controllability. Full A* and simplified A* are detailed in Appendix~\ref{app:astar}. When used with GA, an individual encodes periods and phases of dynamic elements and small ride/wait cost weights that shape the A* cost model.

\paragraph{\textbf{Method 3: TEG-DP}}
Stage~1 treats the dynamic object set and weights as part of the generator parameters $\theta$ (platform/obstacle catalogs, phases, velocities, optional spatial anchors). Given $\theta$, we instantiate $A$ and $F$, and construct per-time forbidden masks and a transition/vertex cost field $c_t(\cdot)\!\ge\!0$. We then run a forward dynamic program on $\widetilde{\mathcal{G}}$ (a layered DAG with edges only from $t$ to $t{+}1$) to obtain a minimum-cost, time-consistent path under $F$:
\[
\textstyle V_{t+1}(x')=\min_{(x\!\to\!x')\in \mathcal{E}_t,\,\neg\text{forbid}_{t+1}(x')}\big\{V_t(x)+c_t(x\!\to\!x')\big\},
\]
with backpointers for path reconstruction and optional constraints. On our finite-horizon layered TEG with nonnegative additive costs, this forward min-cost DP returns the global minimum-cost path; see Appendix~\ref{app:dp} for the argument. We initialize $V_0$ on allowed starts and set $V_t\!=\!\infty$ on forbidden states; the time horizon $T$ prevents cycles. Waiting is encoded by $(x,t)\!\to\!(x,t{+}1)$ with its own cost. 
When $c_t(\cdot)\!\equiv\!1$, the DP degenerates to unweighted shortest path on the time-expanded graph, i.e., layer-by-layer BFS; with structured $c_t$ it becomes a true DP that prefers desired wait/ride patterns and interaction uniformity. The structured field $c_t$ used in all experiments is explicitly defined in Appendix~\ref{app:repro}.
To explore $\theta$ at scale, GA mutates object parameters or cost fields and retains lower-cost candidates via tournament selection, but the planner itself is DP on the TEG. Each candidate is evaluated by the metric vector $m(\pi,A)$, preserving the unified high-dimensional validation of Section~\ref{sec:method:general}.

\begin{table*}[!t]
  \centering
  \small
  \setlength{\tabcolsep}{4.5pt}
  \renewcommand{\arraystretch}{1.05}
  \caption{Large-Scale Experimental Settings for Space Direction.
  Targets use density per 100 path cells; $\lambda$ is the layer-switch cost.}
  \label{tab:space-scales}
  \begin{tabularx}{\textwidth}{l c c c c c c c c X}
    \toprule
    \textbf{Scale} & \textbf{Grid} &
    \textbf{Seeds} & \textbf{GA (pop$\times$gens)} &
    \textbf{$K$} & \textbf{$D_1/D_2$} & $\boldsymbol{\lambda}$ &
    \textbf{Targets} & \textbf{Sweep} & \textbf{Validation settings} \\
    \midrule
    Small  & $30^3$  & $80$ & $40\times30$ & 10 & $2/4$ & 1 &
    $\rho{=}2.0{/}100,\ s_{\min}{=}5$ &
    $\rho\!\in\!\{1,3,5\},\ s_{\min}\!\in\!\{3,5,7\}$ &
    band, $p{=}0.01$, $r_\text{band}{=}1$; ARR-$k{=}10$; stab: pairs$=12$, $|x\!-\!y|_1\!\ge\!20$ \\
    Medium & $50^3$  & $40$ & $ 20\times20$ & 15 & $3/6$ & 1 &
    $\rho{=}2.0{/}100,\ s_{\min}{=}5$ &
    $\rho\!\in\!\{2,4\},\ s_{\min}\!\in\!\{4,6\}$ &
    band, $p{=}0.01$, $r_\text{band}{=}1$; ARR-$k{=}10$; stab: pairs$=10$, $|x\!-\!y|_1\!\ge\!25$ \\
    Large  & $100^3$ & $20$ & $ 10\times10$ & 30 & $4/8$ & 2 &
    $\rho{=}2.0{/}100,\ s_{\min}{=}5$ &
    (no full sweep) &
    global, $p{=}0.005$, $r_\text{band}{=}1$; ARR-$k{=}8$; stab: pairs$=12$, $|x\!-\!y|_1\!\ge\!30$ \\
    \bottomrule
  \end{tabularx}
  \vspace{2pt}
  \raggedright\footnotesize
  \textbf{Notes.} $K$: planned switches; $D_1/D_2$: corridor/room growth depths.
  Weighted-score weights (all scales):
  \texttt{path\_length}: 0.5,\ \texttt{pb\_length\_uniformity}: 1.0,\ \texttt{pb\_spatial\_dispersion}: 2.0,\ \texttt{pb\_min\_length\_penalty}: 2.0,\ \texttt{sb\_coverage}: 1.5.
  PF-A* reward weight: 200 (S/M), 300 (L). Per scale, total runs = \#seeds $\times$ |\#target pairs| $\times$ 3 methods $\times$ 2 modes (single+GA).
\end{table*}

\section{Experiments}
\label{sec:experiments}

This section describes the experimental setups in both directions, including tasks, baselines, and evaluation metrics.

\subsection{Experiment Setup}
\label{sec:exp-setup-space}

\subsubsection{Direction-S}

We evaluate the Space direction on three scales (Small/Medium/Large) following Table~\ref{tab:space-scales}. Each scale instantiates identical planner variants, NP-A*, PF-A*, and NNB, in two modes: \emph{single} (one-shot planning) and \emph{GA} (genetic post-refinement with $G$ generations). For controllability, each scale runs sweeps over target switch density $\rho$ (per 100 path cells) and target minimum spacing $s_{\min}$; the exact sweep sets per scale appear in Table~\ref{tab:space-scales}. Robustness is tested under two perturbation protocols: (i) \textbf{band}-only voxels inside a radius-$r$ tube around the nominal path are randomly blocked with probability $p$; and (ii) \textbf{global}-light, uniform perturbation to the full grid (used on Large). The protocols and parameters (band radius, blocking probability, trial counts) follow Table~\ref{tab:space-scales} and our description of ``band/global micro-perturbations'' used to probe replanning success and cost deltas. 

To evaluate path endpoint sensitivity we also create endpoint pairs per scale with a Manhattan-distance floor and replan after shifting the start/end; near-ceiling success in this probe is used to attribute robustness failures primarily to local micro-structure. For each $(\text{scale}, \text{method}, \text{mode})$ and each target set $(\rho, s_{\min})$, we run the specified number of seeds (Table~\ref{tab:space-scales}) and log all per-run metrics.

\subsubsection{Direction-T}
\label{sec:exp-setup-time}

We evaluate the Time direction on three scales (Small/Medium/Large) with periodic moving platforms and obstacles. Each condition instantiates three methods, Static Backbone, TEG-A*, and TEG-DP, in two modes: single and GA. Per scale we sweep two designer-facing targets: the ride ratio (fraction of frames on platforms) and the minimum inter-event spacing (board/interaction gap in ticks). 

For each $(\text{scale},\text{method},\text{mode})$ and each target pair $(r,d_{\min})$, we run the stated number of seeds (Table~\ref{tab:time-scales}) and log per-run metrics. 
In GA mode, TEG-DP uses a TEG-aware fitness (Section~\ref{sec:method:time}) with population $20$, generations $20$, mutation rate $0.2$, and lightweight evaluators that reuse the same feasibility rules as the validator. 
Plots and statistics are produced by the shared analysis utilities, aggregated by target, method and scale. The reproducibility for both directions, including complete seeds, parameter sweeps, perturbation protocols, GA configurations, and metrics, is documented in Appendix~\ref{app:repro}.

\subsection{Evaluation Metrics and Statistics}
\label{sec:metrics-space}

\subsubsection{Direction-S}

In the Space direction, we evaluate planners with a unified battery of structure, controllability, robustness, and efficiency metrics, and report aggregate statistics per target, method, and scale.

\paragraph{Controllability:} We directly compare targets to achieved structure. 
(1) Density controllability: for each run we estimate the achieved switch density $\hat{\rho}$ (per 100 cells) and visualize $y{=}\hat{\rho}$ against $x{=}\rho$ with a $y{=}x$ reference. 
We also compute mean absolute error (MAE) $(\rho){=}\tfrac{1}{N}\sum_i|\hat{\rho}_i-\rho_i|$ and (when targets have $\ge2$ levels) Spearman $\rho_s$ between $(\rho_i,\hat{\rho}_i)$ for compact tabular summaries.
(2) Spacing controllability (SPC): we turn violation rate into a success rate, 
\[
\text{SPC} \;=\; 1 - \frac{\#\{(u\!\to\!v)\ \text{consecutive switches}: \ \mathrm{dist}(u,v) < s_{\min}\}}{\#\ \text{consecutive switch pairs}},
\]
and aggregate by $s_{\min}$ as mean$\pm$std / 95\% CI per method and scale.

\paragraph{Robustness:} Under micro-perturbations we report the robust replanning success rate (fraction of trials that still find a valid route) and the mean cost delta $\Delta\!\text{cost}$ between perturbed and nominal runs (to characterize the efficiency trade behind robustness). They also summarize the scale-wise behavior and the cost-robustness trade we reproduce in our figures. Endpoint-perturbation success is summarized separately and used diagnostically (near ceiling for all methods/scales). The analysis script provides the endpoint-success plotting with CIs and sample counts per $(\text{scale},\text{method},\text{mode})$.

\paragraph{Alternative Route Robustness:} ARR measures the availability/quality of alternative corridors near the nominal route \cite{eppstein1998finding}; we estimate it per run and visualize distributions by method and scale. ARR exhibits a consistent method ranking and is used to interpret why spacing-compliant methods are also more locally recoverable.

\paragraph{Composite space quality:} We summarize overall quality
\[
\textstyle \mathcal{W} \;=\; \alpha_\ell\,\mathrm{Len} + \alpha_u\,U + \alpha_d\,D + \alpha_m\,M + \alpha_c\,C,
\]
where $\mathrm{Len}$ is path length (lower is better), $U$ is per-block length uniformity, $D$ is spatial dispersion, $M$ is minimum-length penalty, and $C$ is spatial-block coverage; each component is sign-corrected and standardized as z-scores within the same scale (S/M/L); the fixed weights follow the implementation 
$(\alpha_\ell,\alpha_u,\alpha_d,\allowbreak\alpha_m,\alpha_c)=(0.5,1.0,\allowbreak2.0,2.0,1.5)$.
This weight set is chosen to (i) reflect designer-facing priorities that treat controllability and robustness as first-class, (ii) keep efficiency influential but secondary, and (iii) balance magnitudes so that a one-standard-deviation change in each standardized metric contributes a comparable shift to the composite score.

\begin{table*}[t]
  \centering
  \small
  \setlength{\tabcolsep}{5pt}
  \renewcommand{\arraystretch}{1.05}
  \caption{Large-Scale Experimental Settings for the Time Direction. 
  Each scale runs all methods in \emph{single} and \emph{GA} modes.}
  \label{tab:time-scales}
  \begin{tabularx}{\textwidth}{l c c c c c c X}
    \toprule
    \textbf{Scale} & \textbf{Grid \& Horizon} & \textbf{Seeds} & \textbf{GA (pop$\times$gens)}
    & \textbf{Plats/Obs} & \textbf{Min spans (P/O)} 
    & \textbf{Sweep} & \textbf{Options} \\
    \midrule
    Small  & $30{\times}15;\ T_{\max}{=}200$ & $12$ & $20{\times}20$ 
           & $4/4$  & $4/3$ 
           & $r\in\{0.30,0.40\},\ d_{\min}\in\{10,12\}$ 
           & allow-ride,\ TTC on \\
    Medium & $50{\times}25;\ T_{\max}{=}300$ & $8$ & $20{\times}20$ 
           & $5/5$  & $5/4$ 
           & $r\in\{0.25,0.35\},\ d_{\min}\in\{12,15\}$ 
           & allow-ride,\ TTC on \\
    Large  & $80{\times}40;\ T_{\max}{=}500$ & $4$ & $20{\times}20$ 
           & $8/8$  & $6/5$ 
           & $r\in\{0.30,0.40\},\ d_{\min}\in\{15,18\}$ 
           & allow-ride,\ TTC on \\
    \bottomrule
  \end{tabularx}
  \vspace{2pt}
  \raggedright\footnotesize
  \textbf{Notes.} Plats/Obs and Min spans specify counts and minimal track spans for objects. 
  ``TTC on'' enables time-to-collision (TTC) and head-on swap checks in TEG feasibility.
\end{table*}

\paragraph{Additional structure/efficiency diagnostics:} 
We also report per-run or per-target distributions for: (i) dead-end rate (DER), (ii) switch-exit freedom (SEF; local openness around switch portals), (iii) tortuosity, (iv) branching factor, (v) layer balance, and (vi) wall-clock time. These are summarized as figure panels and/or columns in the aggregated CSVs and are used primarily to interpret the main findings; see section~\ref{app:space-more} for how SPC/SEF/ARR complement the controllability and robustness story.

\paragraph{Aggregation and uncertainty:} Unless otherwise stated we aggregate over seeds at the (target, method, scale) level and report mean with either standard deviation or a normal-approximation 95\% CI (code paths for mean/std and CI-based rendering are shown in the controllability and endpoint-success plotting utilities).

\subsubsection{Direction-T}
\label{sec:metrics-time}
For the Time direction we compute all statistics in $\mathcal{O}(T)$ from the validated TEG policy (same feasibility rules as the validator) and aggregate by (target, method, scale) with mean$\pm$std or 95\% CIs. Metrics are organized but reported jointly: (i) Temporal controllability-ride-ratio error $\mathrm{MAE}_r{=}|\,\hat r-r\,|$ (achieved $\hat r$ vs.\ target $r$) and inter-event spacing success defined as $1$ minus the violation rate for the target minimum gap $d_{\min}$; (ii) Pacing-uniformity of interaction beats via the coefficient of variation of inter-event intervals (lower is steadier); (iii) Playability \& safety-waiting ratio (idle frames), transfer count (board/alight events), and near-swap counts with moving obstacles detected along the executed path; (iv) Spatial efficiency-ticks-to-goal (path length) and coverage ratio (distinct visited cells over open cells) plus optional local tortuosity; and (v) a composite utility $J_{\text{time}}$ that linearly combines length/coverage/ride/wait/interaction components (the same weighting used by the GA evaluator) together with runtime (wall-clock seconds). These summaries support the same plots as in Space-target vs.\ achieved curves (for $r$ and $d_{\min}$), distributional pacing/interaction diagnostics, and runtime charts-enabling consistent, cross-direction comparison within the HDPCG framework.

\subsection{Unity Case Study Setup}

\subsubsection{Direction-S}
We build a Unity pipeline to instantiate the \emph{Space} direction levels produced by the planner. The analysis code exports a compact JSON per level with grid size $(W,H)$ (or $(W,H,D)$), start/end, and two binary occupancies $\{L_0,L_1\}$ (0: free, 1: solid). A LevelManager loads this file and constructs two aligned voxel meshes with colliders. Both layers are always resident; switching never reloads assets. The player uses a light-weight controller (WASD), and reaching the end trigger completes the level. This Unity scene supports direct human play; however, all quantitative results in Sec.~\ref{sec:results-space} are computed from solver-generated trajectories rather than from user studies.

We support two switching realizations that correspond one-to-one to our planner's semantics. For 2.5D gravity inversion, pressing the switch key negates the local gravity vector and swaps which layer is treated as ground: floors/ceilings exchange roles while switch pockets from the plan remain traversable. For 3.5D time-shift, a toggle key enables rendering+colliders on the active timeline and disables them on the inactive one; an optional time lens renders the inactive layer through an elliptical mask for preview before committing to a switch. These mechanics are implemented with a thin set of scripts, require no hand-authored stitching, and operate directly on the exported planner grids. The two Unity scenes used in Section~\ref{sec:results-qual-space} are built exactly from the same JSONs that produce our Python visualizations, ensuring that controllable switch density and spacing measured in Section~\ref{sec:results-space} manifest identically in-engine.

\subsubsection{Direction-T}
Similarly, we instantiate Time-direction levels in Unity using the same high-level specification as our planners. Each level description includes the grid bounds, static tiles, a catalog of dynamic elements (platforms and obstacles) with waypoint polylines, periods and phases, and the validated player policy as a time-indexed sequence of primitives \{WAIT, WALK, RIDE\}. At load time the scene is procedurally built: static geometry is meshed; each dynamic element is bound to a parametric mover that replays its trajectory with the prescribed timing; and a lightweight controller replays the exported policy. The controller enforces the same feasibility rules as the TEG: waiting in place, collision-free walking to the next tick, boarding/alighting only at platform endpoints, and the prevention of head-on swaps with moving obstacles. A single clock drives movers and controller to guarantee tick-accurate synchronization with the planner's discretization.

For qualitative analysis we render the player path as a ribbon with thicker strokes on ride segments and small decals at wait nodes, and color-code dynamic tracks (platforms in green, obstacles in red) to match the Python structural view. The loader can display either a one-shot plan single or the best candidate selected by evolutionary search GA; both are validated on the same TEG rules, so the Unity replay is a faithful realization of the computed solution. This setup validates that our temporal modeling supports a wide range of modern platforming mechanics while remaining controllable by designer targets such as ride ratio and minimum inter-event spacing, enabling direct comparison between quantitative metrics and experiential pacing in a real-time 3D environment.

\section{Results}
\label{sec:results}

This section presents the quantitative and qualitative results of these experiments, showing how HDPCG methods compare to baselines and what design-relevant patterns emerge.

\subsection{Space: Quantitative Results}
\label{sec:results-space}

\paragraph{Setup recap.}
We evaluate three planners within the HDPCG \emph{Space} direction( NP-A*, PF-A*, and NNB) across three scales \{S,M,L\}. Unless noted, we report single-run statistics aggregated over the large experiment suite; full results are deferred to Appendix~\ref{app:space-more}.

\paragraph{R1 - Overall space quality:}

\begin{figure}[t]
  \centering
  \includegraphics[width=\linewidth]{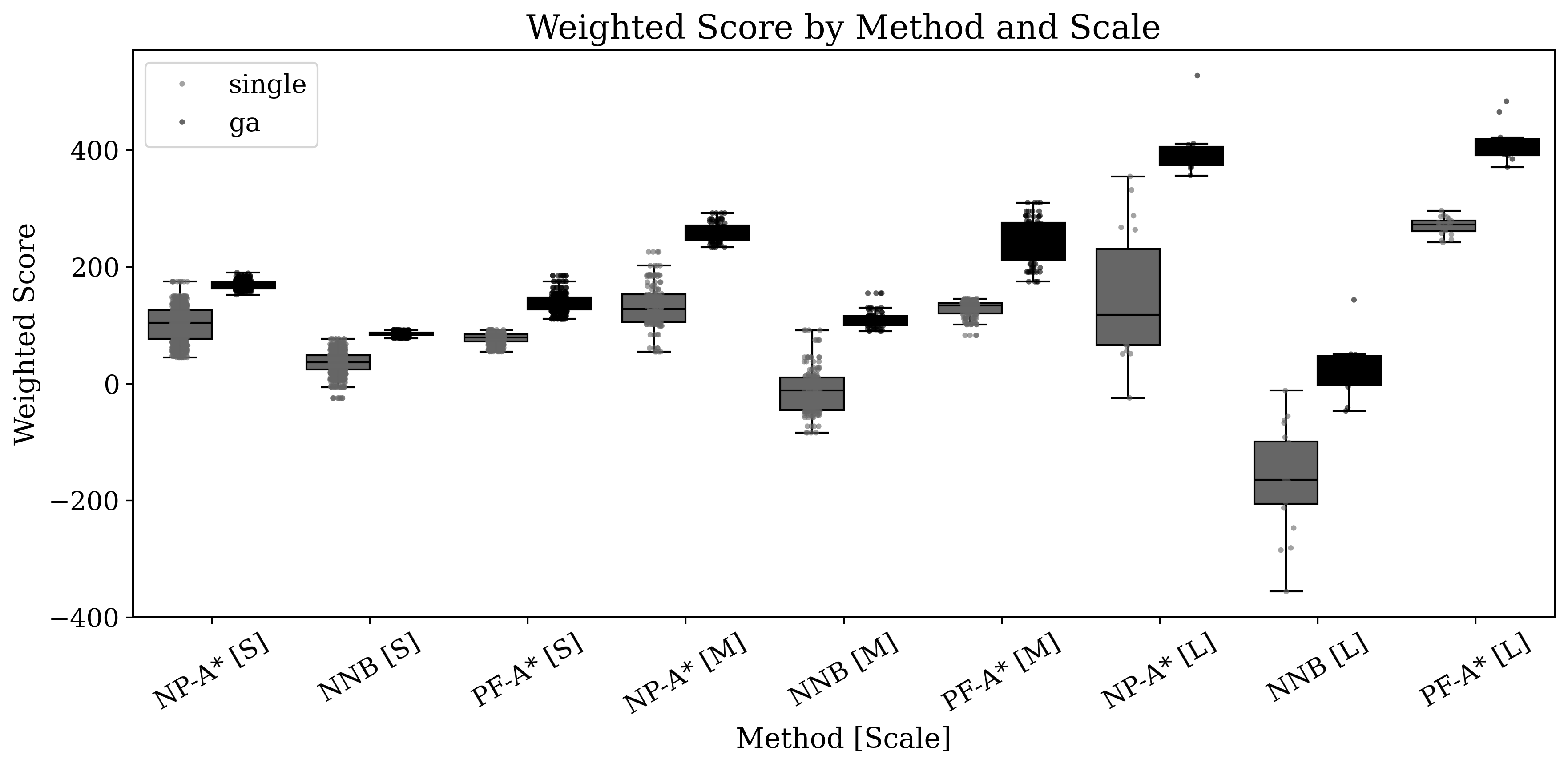}
  \caption{\textbf{Space quality.} S/M: NP-A* $>$ PF-A* $>$ NNB; L: PF-A* $>$ NP-A* $\gg$ NNB. GA boosts all methods; PF-A* best at L.}
  \label{fig:ws}
\end{figure}

Fig.~\ref{fig:ws} compares the composite Weighted Score. On S/M, NP-A* attains the highest means (e.g., S: 168.9 vs.\ 137.6 for PF-A* in GA; M: 259.6 vs.\ 241.5), while NNB trails. On L, PF-A* overtakes (GA means: 412.3 for PF-A* vs.\ 400.0 for NP-A*; single: 99.6 vs.\ 45.4), with NNB collapsing (single mean $-327.5$). This establishes a scale-dependent ranking: NP-A* dominates at small/medium sizes; PF-A* scales better and wins at large size. Because W aggregates z-scored components, absolute values (including negatives) reflect relative performance within scale; the ordering is the intended interpretation.

\paragraph{R2 - Controllability of switching patterns:}

\begin{figure}[t]
  \centering
  \includegraphics[width=\linewidth]{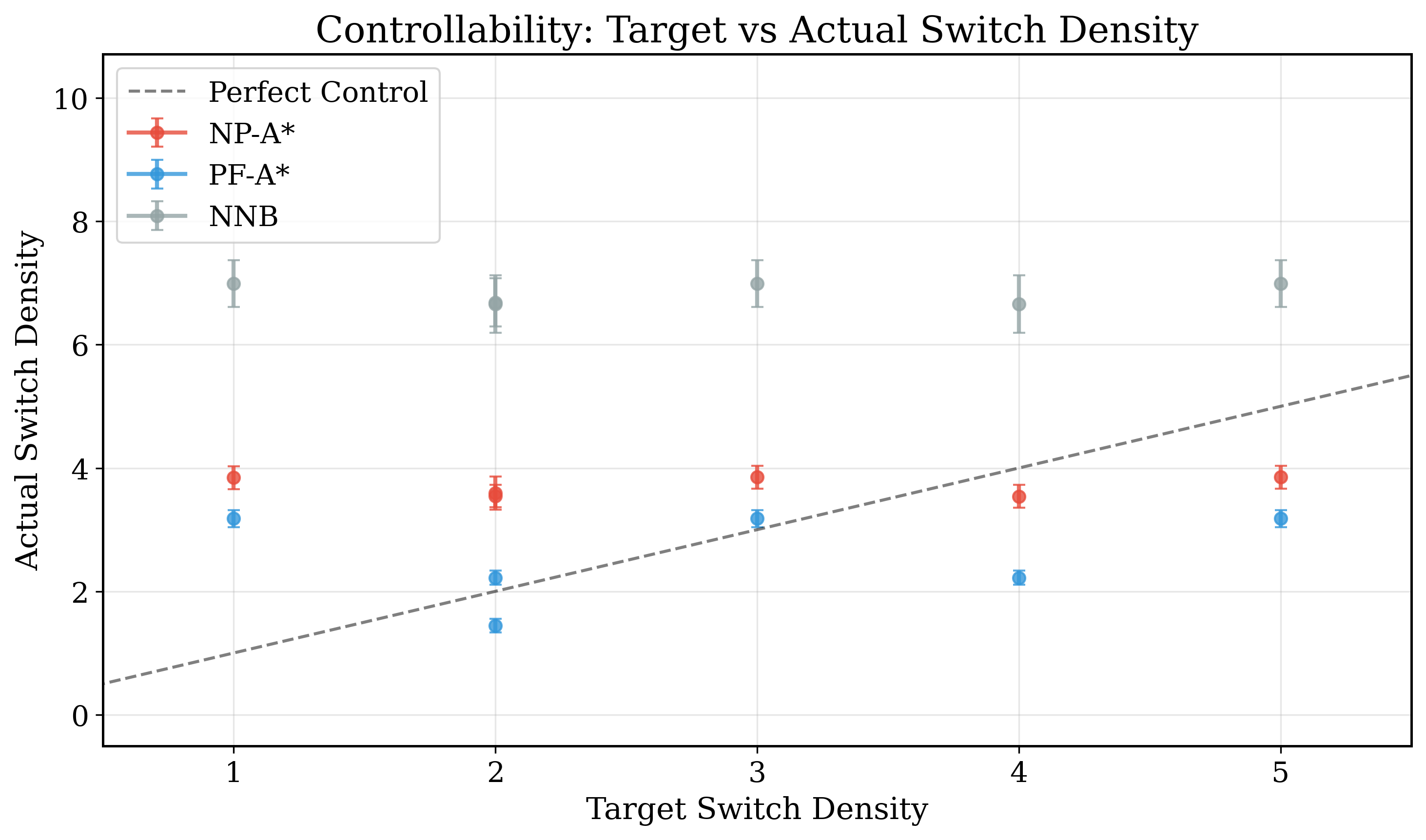}
  \caption{\textbf{Controllability-density.} PF-A* tracks density targets with lowest MAE at L and competitive at S/M; NP-A* exhibits mild systematic overshoot; NNB is insensitive to targets.}
  \label{fig:ctrl-density}
\end{figure}

\begin{figure}[t]
  \centering
  \includegraphics[width=\linewidth]{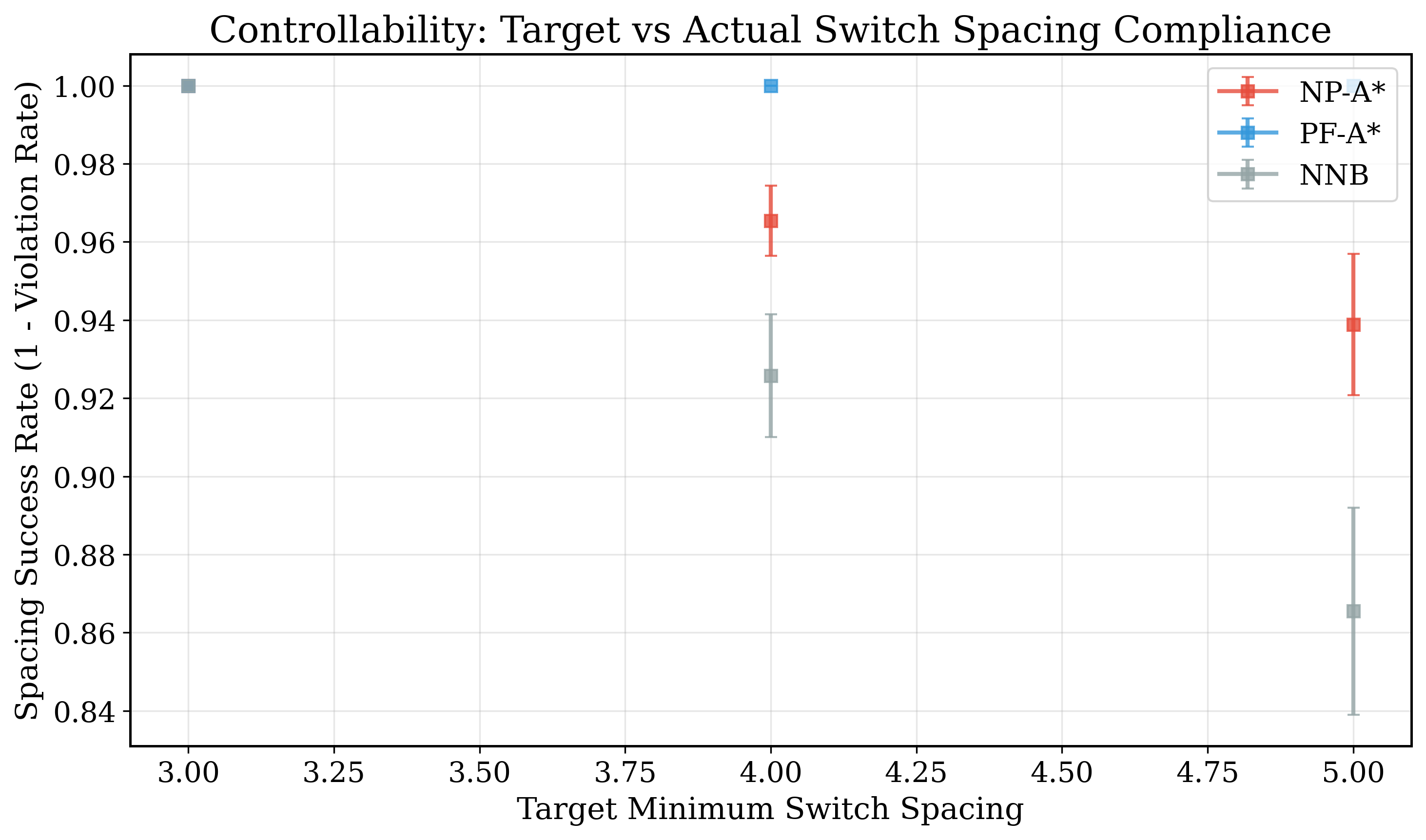}
  \caption{\textbf{Controllability-spacing.} PF-A* $\approx$ perfect across scales; NP-A* slightly lower; NNB drops with higher targets.}
  \label{fig:ctrl-spacing}
\end{figure}

We test whether target switch density and minimum spacing materialize in the final geometry. Fig.~\ref{fig:ctrl-density} and Fig.~\ref{fig:ctrl-spacing} visualize target vs.\ achieved values; the aggregated table (\autoref{tab:controllability_summary}) reports error statistics. Spacing control: PF-A* is near-perfect across scales (mean MAE $\approx$ 0.00 at S/M/L), NP-A* is close but nonzero ($\approx$ 0.09), and NNB degrades with larger spacing targets ($\approx$ 0.28-0.33). Density control: at M, NP-A* shows slightly lower MAE ($\approx$1.37) than PF-A* ($\approx$2.08), while at L PF-A* is clearly better ($\approx$1.34 vs.\ 2.55); at S the two are similar ($\approx$2.22 vs.\ 2.25). Overall, PF-A* provides the most reliable knob-to-behavior mapping, especially for spacing, with NP-A* trading mild density overshoot for higher composite score (R1).

\begin{table}[t]
  \centering
  \small
  \setlength{\tabcolsep}{5.5pt}
  \caption{\textbf{Controllability summary (MAE $\downarrow$).} 
  Mean absolute error between target and achieved \emph{Density} / \emph{Spacing} aggregated per scale (S/M/L). 
  Best per column is \textbf{bold}. 
  Counts (runs aggregated per scale): S: $1080$; M: $240$; L: $30$}

  \label{tab:controllability_summary}
  \begin{tabular}{lcccccc}
    \toprule
    & \multicolumn{2}{c}{\textbf{S}} & \multicolumn{2}{c}{\textbf{M}} & \multicolumn{2}{c}{\textbf{L}} \\
    \cmidrule(lr){2-3}\cmidrule(lr){4-5}\cmidrule(lr){6-7}
    \textbf{Method} & \textbf{Den.} & \textbf{Sp.} & \textbf{Den.} & \textbf{Sp.} & \textbf{Den.} & \textbf{Sp.} \\
    \midrule
    PF-A*  & 2.247 & \textbf{0.002} & 2.078 & \textbf{0.000} & \textbf{1.343} & \textbf{0.000} \\
    NP-A*  & \textbf{2.217} & 0.091 & \textbf{1.366} & 0.093 & 2.554 & 0.097 \\
    NNB    & 9.410 & 0.288 & 8.738 & 0.284 & 10.978 & 0.326 \\
    \bottomrule
  \end{tabular}
\end{table}

\paragraph{R3 - Robustness under perturbations:}
Under stochastic voxel micro-perturbations (band/global protocols as in Section~\ref{sec:experiments}), \autoref{fig:robust-rate} shows re-planning success by scale. Because the GA fitness maximizes $W$ without a robustness term, it preferentially prunes the ``slack'' and redundant corridors that make PF-A* resilient, producing higher-$W$ but brittle, low-redundancy layouts with lower perturbation success (cf.\ ARR in Appendix~\ref{app:space-more}). Only PF-A* (single) achieves modest robustness: S $0.076\pm0.26$, M $0.045\pm0.208$, L $0.217\pm0.40$; NP-A* and NNB are near zero across scales, and GA variants are also near zero. This suggests PF-A*'s strong spacing regularity (R2) translates into better local recoverability.

\paragraph{R4 - Efficiency-coverage trade-off:}
PF-A* populates a favorable frontier (high coverage at shorter lengths); NP-A* can push coverage to the maximum but often at longer paths; NNB is dominated. This explains the interplay between R1 and R2-R3: NP-A* wins the composite objective on S/M, while PF-A* offers a cleaner length-coverage trade and stronger controllability/robustness at L. Full details in Appendix~\ref{app:space-more}.

\paragraph{R5 - Runtime:}
We report wall-clock time (mean~$\pm$~std; seconds) for both single and GA modes in Table~\ref{tab:runtime-all}. 
Two consistent patterns emerge: (i) in single mode the naive baseline (NNB) is always fastest across S/M/L, reflecting its minimal shaping; 
(ii) in GA mode, NNB remains fastest on S/M, but at L the segment-wise shaping of NP-A* has the lowest runtime (199.19~s), while PF-A* is slowest (500.06~s) yet achieves the best large-scale quality (R1). 

\begin{table}[t]
  \centering
  \small
  \setlength{\tabcolsep}{5pt}
  \caption{\textbf{Runtime (seconds; lower is better; Mean~$\pm$~std.)}}
  \label{tab:runtime-all}
  \begin{tabular}{lccc}
    \toprule
    \textbf{Method (mode)} & \textbf{S} & \textbf{M} & \textbf{L} \\
    \midrule
    PF-A* (single) & $0.138 \pm 0.060$ & $0.663 \pm 0.278$ & $4.119 \pm 1.136$ \\
    NP-A* (single) & $0.248 \pm 0.094$ & $0.982 \pm 0.419$ & $3.542 \pm 0.915$ \\
    NNB\ \ (single) & $\mathbf{0.070 \pm 0.032}$ & $\mathbf{0.366 \pm 0.173}$ & $\mathbf{3.163 \pm 1.597}$ \\
    \midrule
    PF-A* (GA)     & $35.888 \pm 6.045$  & $186.431 \pm 26.924$ & $500.064 \pm 37.407$ \\
    NP-A* (GA)     & $43.172 \pm 7.086$  & $153.150 \pm 9.749$  & $\mathbf{199.189 \pm 9.823}$ \\
    NNB\ \ (GA)     & $\mathbf{25.368 \pm 4.292}$ & $\mathbf{119.232 \pm 15.990}$ & $411.569 \pm 44.407$ \\
    \bottomrule
  \end{tabular}
\end{table}

Across scales, (i) NP-A* delivers the best composite score on S/M, while PF-A* scales best and wins at L; 
(ii) PF-A* offers the strongest knob-to-behavior controllability (near-perfect spacing) and the highest robustness under micro-perturbations; 
(iii) NNB is a useful lower-bound baseline and the fastest in single mode, whereas NP-A* is the most GA-efficient at L.

\subsection{Space: Qualitative Analysis + Case Study}
\label{sec:results-qual-space}

\begin{figure}[t]
  \centering
  \includegraphics[width=.9\linewidth]{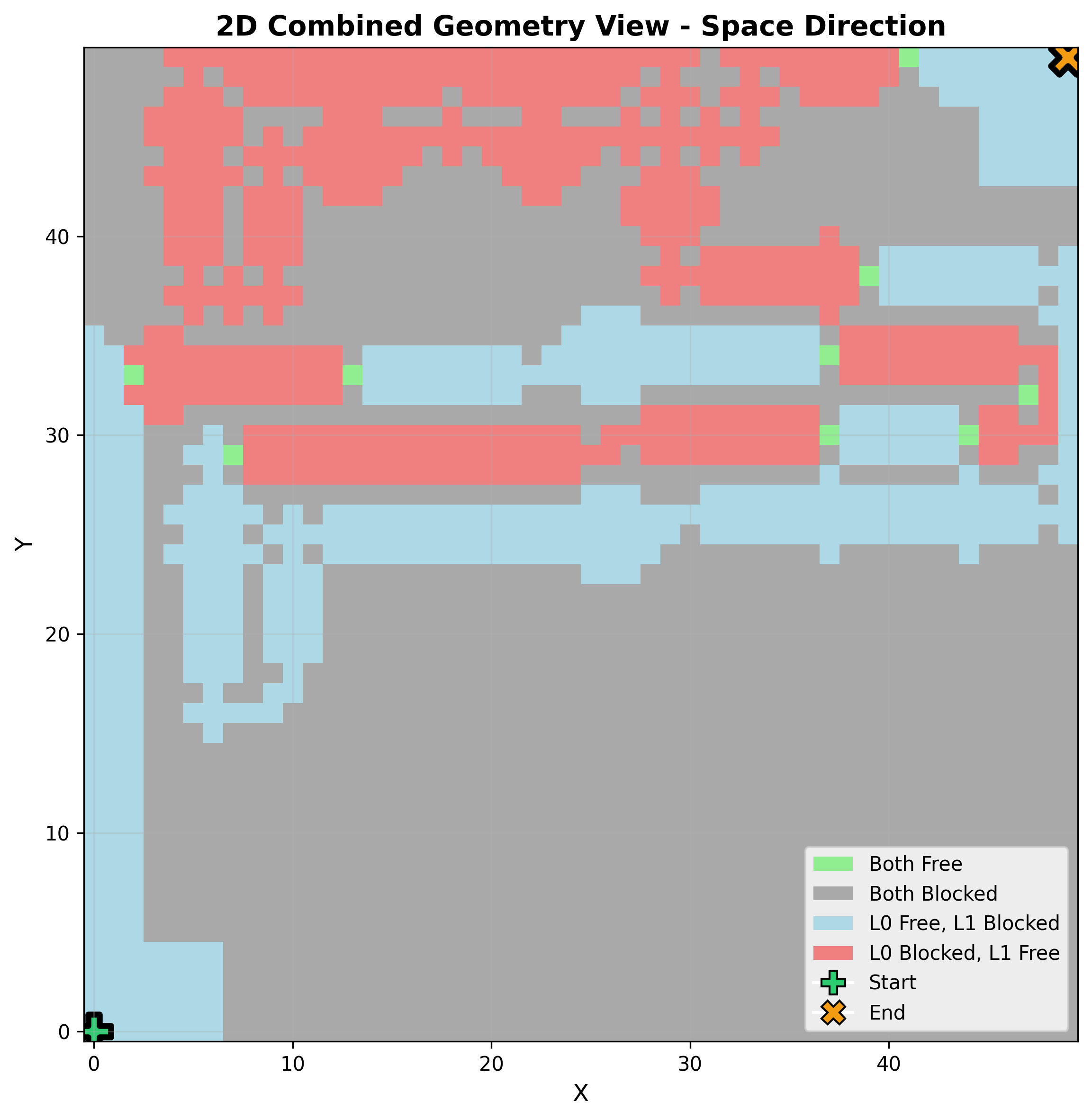}
  \caption{\textbf{2D combined-geometry view (Space).}
  Blue/red cells are free only on $L_0/L_1$; green are free on both (switch pockets); grey are blocked on both.
  Start/end are overlaid. This top-down grid visualization corresponds to our 2.5D Space (gravity-shift) instantiation. The corridor layout and pocket placement obey the target \emph{density}/\emph{spacing} used in Sec.~\ref{sec:results-space}.}
  \label{fig:space-2d-combined}
\end{figure}

\begin{figure}[t]
  \centering
  \includegraphics[width=.9\linewidth]{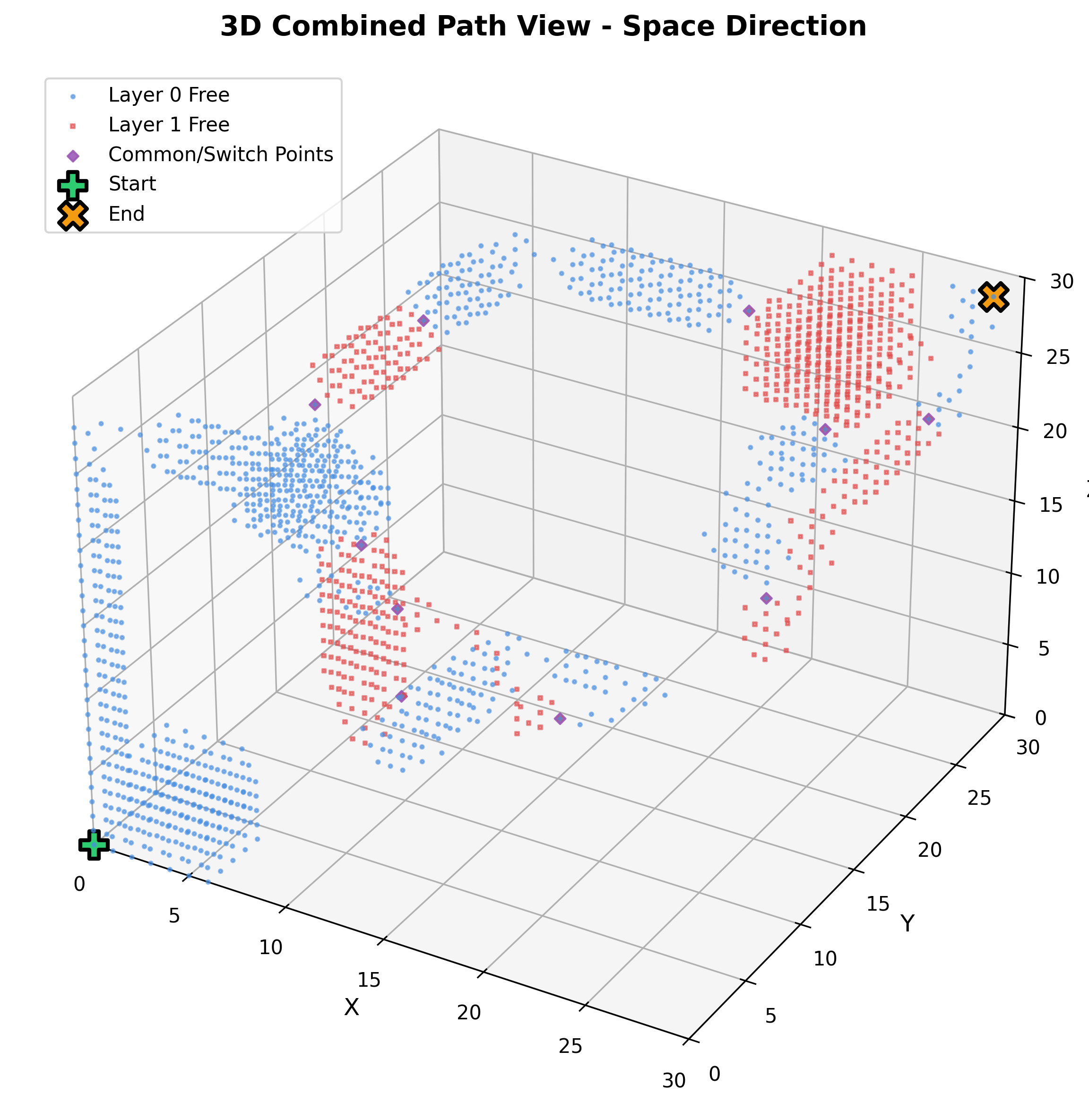}
  \caption{\textbf{3D combined-path view (Space).}
  Blue/red points are free voxels on $L_0/L_1$; purple diamonds are planned switch nodes on the successful route. Alternation frequency matches the requested spacing.}
  \label{fig:space-3d-combined}
\end{figure}

\begin{figure}[t]
  \centering
  \includegraphics[width=\linewidth]{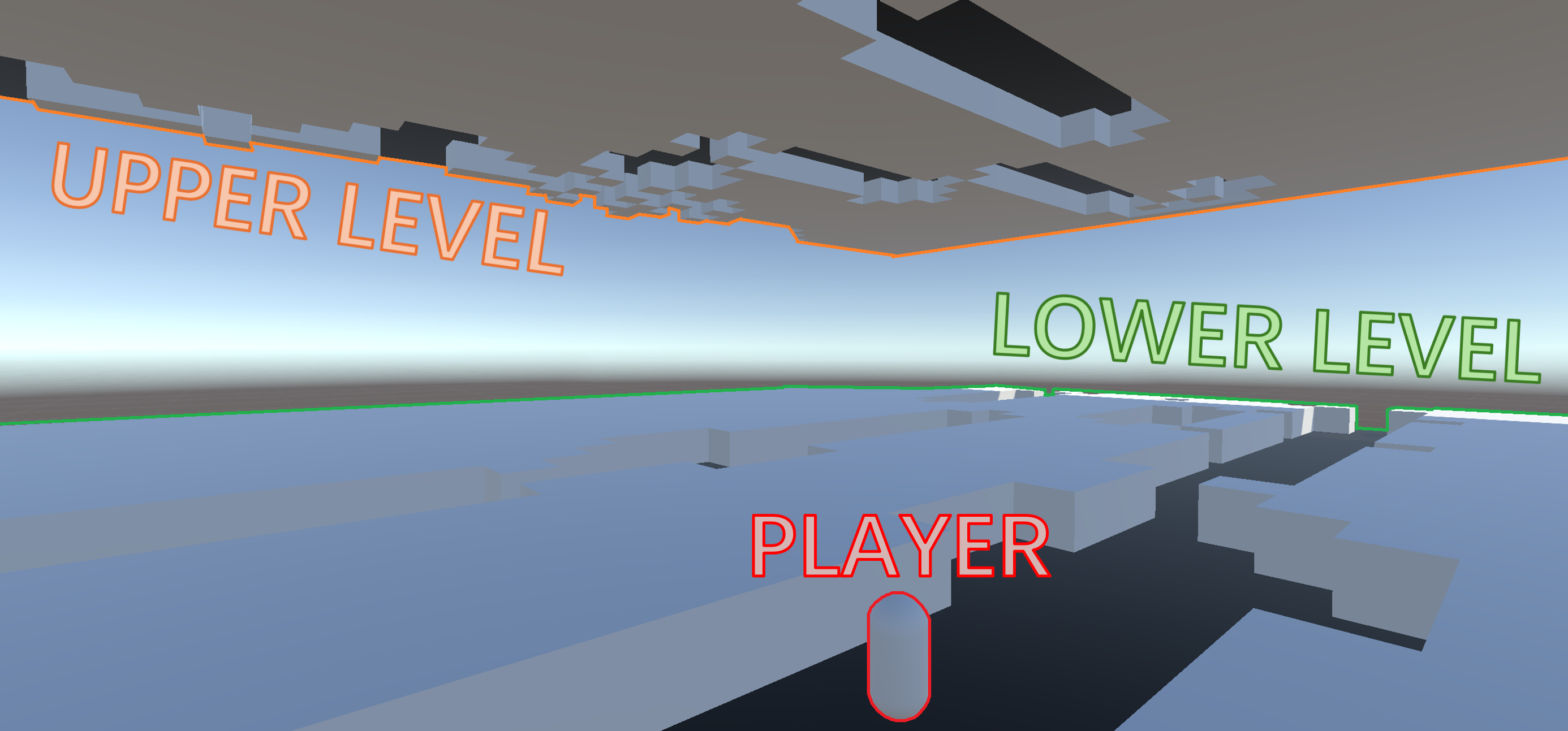}
  \caption{\textbf{Unity snapshot (2.5D gravity inversion).}
  Flipping gravity swaps the active walkable layer; corridors and apertures correspond one-to-one with Fig.~\ref{fig:space-2d-combined}.}
  \label{fig:space-unity-2d}
\end{figure}

\begin{figure}[t]
  \centering
  \includegraphics[width=\linewidth]{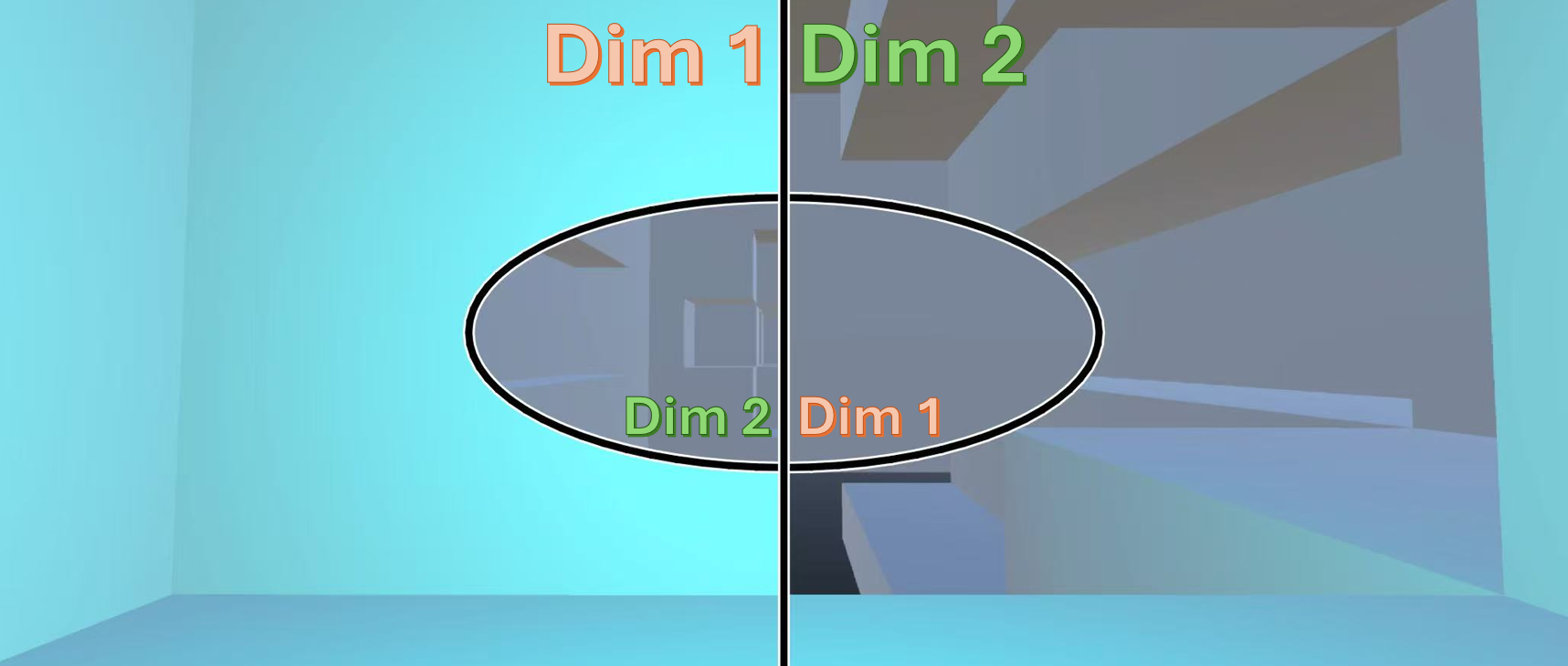}
  \caption{\textbf{Unity snapshot (3.5D time-shift).}
  The \emph{lens} previews the alternate timeline; toggling swaps active/inactive layers. Planned switch points remain reachable post-toggle.}
  \label{fig:space-unity-3d}
\end{figure}

We showcase two representative levels produced by the \emph{Space} planner and instantiated in Unity, chosen to mirror well-known mechanics in real games: a \textbf{2.5D gravity-inversion} level in the spirit of gravity-flip platformers (e.g., \emph{VVVVVV}), and a \textbf{3.5D time-shift} level inspired by time-toggle encounters in \emph{Dishonored 2: A Crack in the Slab} and \emph{Titanfall 2: Effect and Cause}. The Python views in Fig.~\ref{fig:space-2d-combined}-\ref{fig:space-3d-combined} render the exact grids consumed by Unity; Figs.~\ref{fig:space-unity-2d}-\ref{fig:space-unity-3d} are in-engine screenshots of those same levels.

\begin{figure*}[t]
  \centering
  \includegraphics[width=\linewidth]{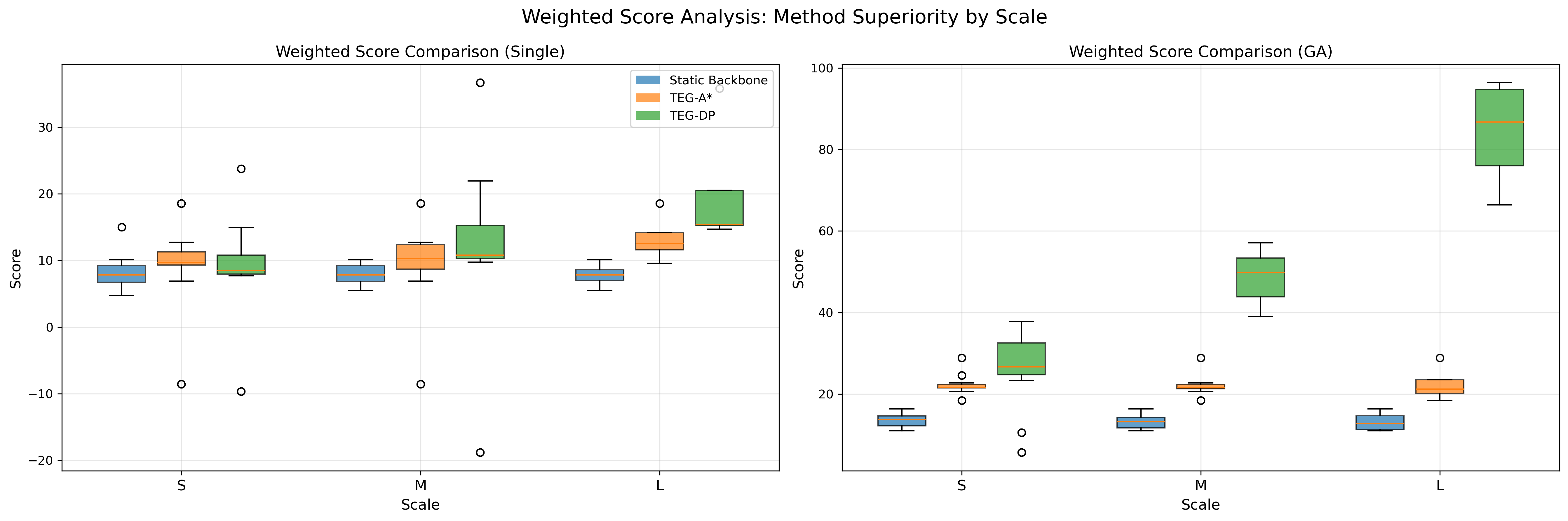}
  \caption{\textbf{Time (scores).} Weighted score distributions across scales and modes.
  The ordering is consistent: TEG-DP $>$ TEG-A* $>$ Static, and GA pushes medians higher, especially at Scale L.}
  \label{fig:time_scores_fig}
\end{figure*}

\textbf{Case A - 2.5D gravity inversion.} 
Fig.~\ref{fig:space-unity-2d} shows two co-located slab layers with mirrored walkable surfaces. A single toggle flips the gravity vector and swaps which layer acts as \emph{ground}; the player ``falls'' through aligned apertures at green pockets (Fig.~\ref{fig:space-2d-combined}) to continue. Because switch pocket density and minimum spacing are set by our planner (Sec.~\ref{sec:results-space}, R2), designers can dial frequent micro-switch beats for dexterity tests, longer runs for traversal pacing-without any hand stitching. The Unity level is generated directly from the planner; what you see is exactly what is playable.

\textbf{Case B - 3.5D time-shift with preview lens.}
In Fig.~\ref{fig:space-unity-3d}, an elliptical time lens reveals the inactive timeline while the player stands in the active one; a toggle swaps colliders/materials so that only one layer is traversable at a time. This reproduces the read\,$\rightarrow$\,plan\,$\rightarrow$\,toggle\,$\rightarrow$\,traverse beat common to modern time-shift encounters, while preserving the planned switch nodes (Fig.~\ref{fig:space-3d-combined}) and their spacing for fairness. The same structural knobs (density/spacing) govern the cadence of time jumps, enabling both rapid multi-hop sequences and slower, set-piece transitions.

Across both cases, the same two-layer abstraction supports complex mechanics, gravity inversion and time shift, because our planner explicitly models layer-specific free space plus shared switch pockets. The quantitative controls measured in Sec.~\ref{sec:results-space} (density / spacing, robustness pockets) translate into qualitative pacing and readability: designers can target rhythmic alternation, guaranteed clearance at pockets, and route regularity. Combined with the automatic Unity instantiation, these examples show practical applicability: HDPCG's Space direction can block out modern switching-based levels with controllable structure and beat-level pacing, aligning with patterns seen in shipped games.

\subsection{Time: Quantitative Results}
\label{sec:exp:time}

We compare the three Time-direction validators, Static Backbone, TEG-A* (simplified TEG search), and TEG-DP, under both Single and GA across scales S/M/L. The weighted score aggregates playability/interaction metrics; higher is better. Table~\ref{tab:time-score-combined} reports the means and standard deviations; Fig.~\ref{fig:time_scores_fig} visualizes the score distributions. Additional results are in Appendix~\ref{app:time}.

Overall, the weighted score generally follows the order of TEG-DP $>$ TEG-A* $>$ Static.
This ordering is statistically significant after Holm-Bonferroni correction in most cases (except S-Single and M-Single); see Appendix~\ref{app:time}, Table~\ref{tab:significance} for $U$ statistics, Cliff's \(\delta\), and adjusted \(p\)-values. GA further amplifies this gap, especially at L. Runtime shows the expected cost-quality trade-off: DP delivers the best playability/interaction patterns at a higher cost; A* offers an interpretable and much faster compromise; the Static validator is a lightweight reference for ablations.

\begin{figure}[t]
  \centering
  \includegraphics[width=\linewidth]{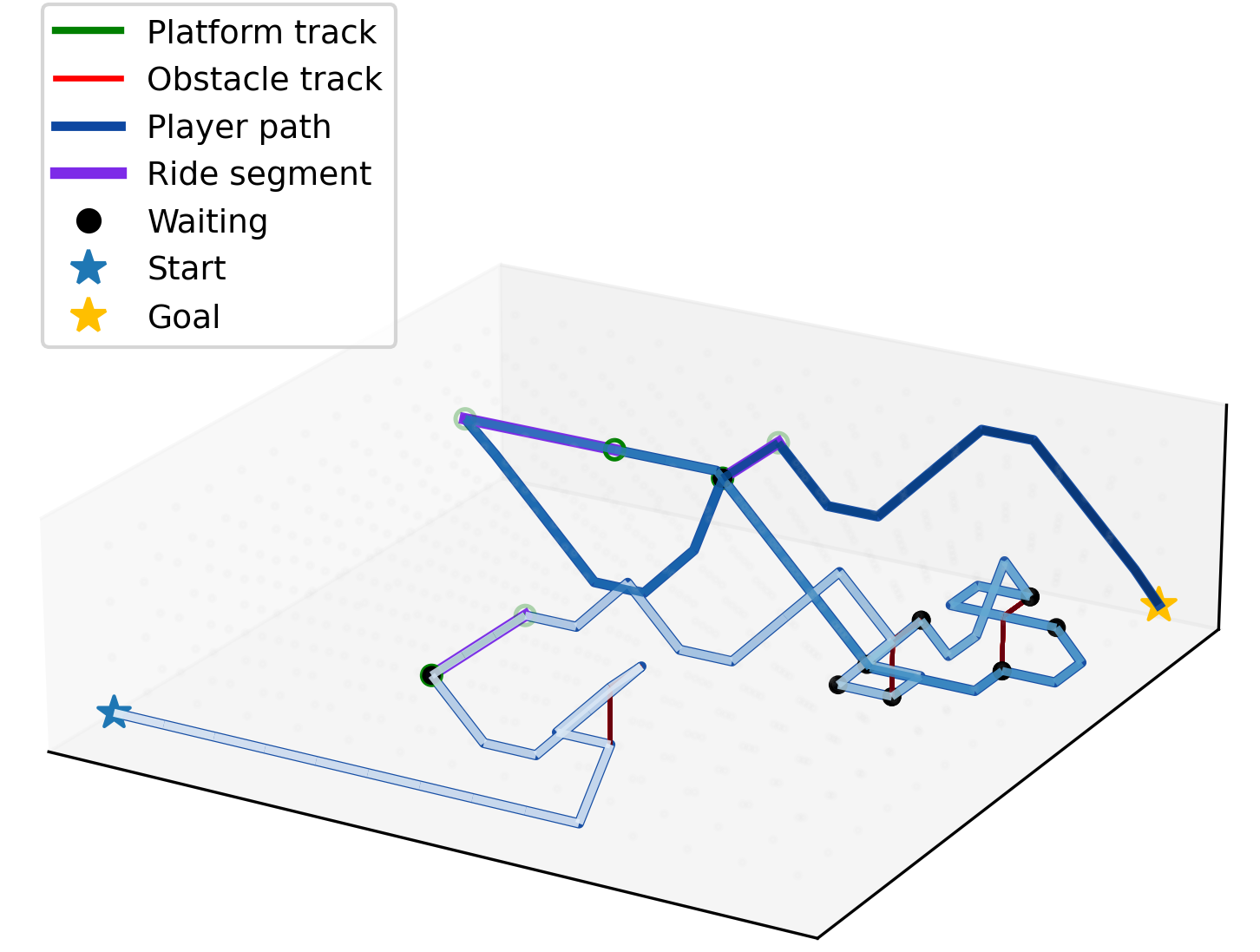}
  \caption{Time direction (Python view). Green: platform tracks; red: obstacle tracks; blue: validated player path (darker $\rightarrow$ later); purple: ride segments; black points: waits. Stars mark start/goal.}
  \label{fig:time-python-viz}
\end{figure}

\begin{figure}[t]
  \centering
  \includegraphics[width=\linewidth]{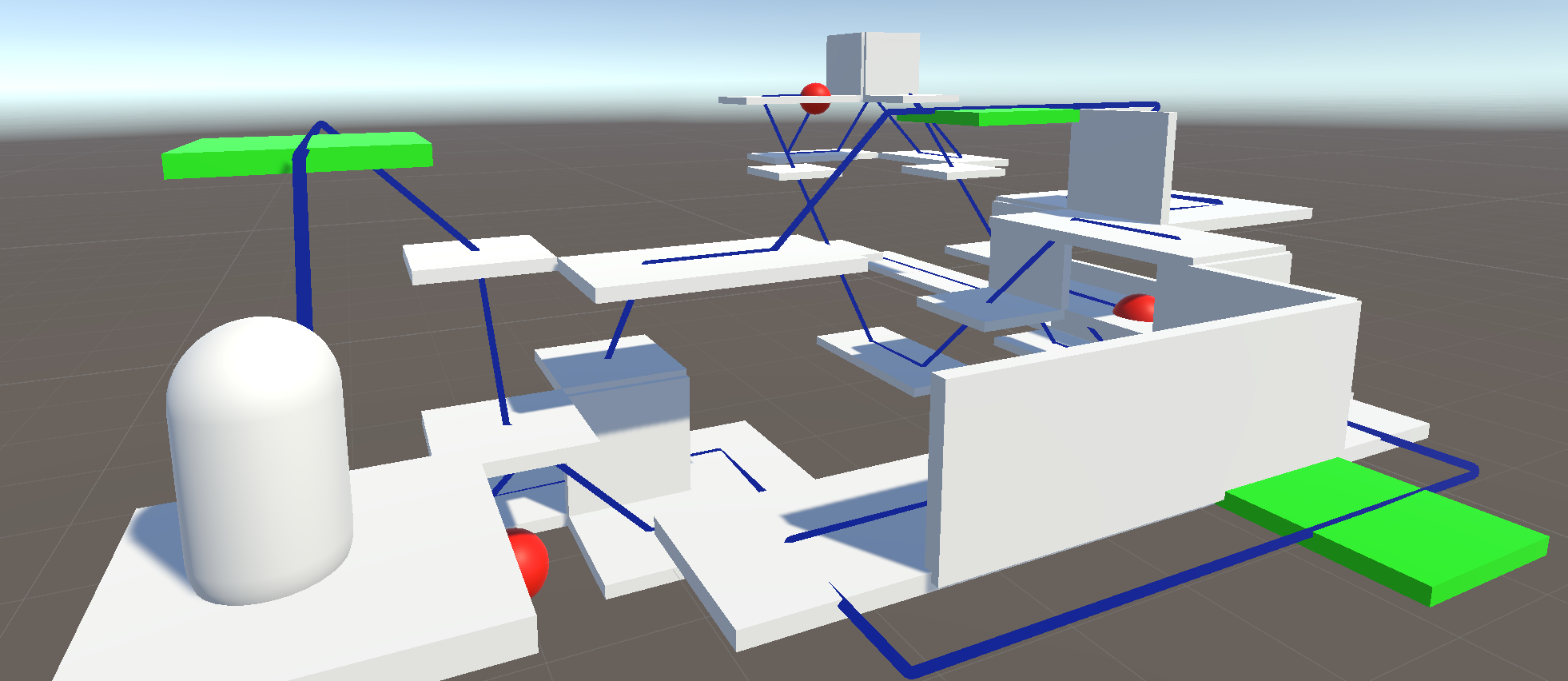}
  \caption{Time direction (Unity snapshot). The 3D platformer level instantiates the same moving platforms (green) / enemies (red) and reproduces the planned route (blue).}
  \label{fig:time-unity-snap}
\end{figure}

\subsection{Time: Qualitative Analysis + Case Study}
Figure~\ref{fig:time-python-viz} shows a single, fully-time-aware level generated by our Time direction and rendered by the Python visualizer; Figure~\ref{fig:time-unity-snap} shows the same instance reconstructed in Unity as a playable 3D platformer in the style of contemporary action-platform games. The Time model uses a time-expanded graph (TEG) to encode \emph{wait}, \emph{walk}, and \emph{ride} transitions with conflict rules that forbid head-on swaps with moving obstacles and allow boarding/alighting only at platform endpoints. The Python figure overlays: (i) platform tracks (green polylines) and obstacle tracks (red polylines); (ii) the validated player trajectory (blue polyline, darker when later in time); (iii) ride segments (purple overlays); and (iv) discrete waiting events (black markers). Start and goal are shown as star glyphs. The trajectory demonstrates our method's ability to interleave purposeful waits with multi-segment rides while respecting safety constraints; the cadence matches target ratios specified to the generator.

In Unity we instantiate identical dynamics: each platform and obstacle follows the exported keyframed path with cycle time and phase; the player can board at the scripted endpoints and is carried between frames while collisions block illegal swaps. Despite being simulated in continuous time, the observed play matches TEG semantics, including the discrete wait/ride structure and the same start-to-goal itinerary. This close match shows that our temporal formulation captures the key elements of moving-platform encounters, while letting designers control pacing through targets such as ride ratio and minimum inter-event spacing.

\section{Conclusion}
\label{sec:conclusion}

In this paper, we introduced High-Dimensional Procedural Content Generation, a graph-based framework that unifies space with explicit mechanic axes, and we show how this representation supports constructive solvers and compilation for mechanics such as layer switching and moving platforms. 
We formalized a general problem statement and a shared four-stage pipeline, then instantiated it in two domains with baselines and structured methods under the same representation. Across multi-scale experiments, our methods achieve controllable targets and replayable solutions, and Unity case studies confirm that the exported plans reproduce the intended mechanics in-engine rather than only matching geometry. Because validation certifies a witness route rather than enumerating all reachable behaviors, the instantiated levels can admit emergent shortcuts after corridor growth, which we leave for future work.

However, several limitations remain. First, our ideal TEG-A* for Time requires state augmentation with interaction memory and rewards to drive meaningful, non-redundant contacts; this causes state explosion, forcing us to use a simplified variant that is tractable but less capable of strictly shaping interactions, leaving a gap between the Static baseline and DP. Second, the representation and solvers we study are constructive/search-based; we do not (yet) integrate PCGML or reinforcement-learning approaches that could learn priors or value functions over the high-dimensional state space. Third, we evaluate each mechanic axis separately; we have not yet evaluated multi-mechanic compositions in a single joint state space or their scalability. Fourth, robustness is currently evaluated post-hoc via re-planning under micro-perturbations, but is not yet part of GA fitness because it would require multiple re-planning rollouts and significantly increase cost. Lastly, we have not conducted user studies; playability and pacing are supported by metrics and replays, but formal evidence of fun is future work.

To address these, we will develop multi-tier planners with bounded-suboptimal search, event-window unrolling, and stronger pruning to make richer A* feasible at interactive scales.
Although we instantiate HDPCG on 2D/3D grids here, the formulation only assumes a discrete adjacency graph; extending it to other discretizations is a natural direction for future work. Because the witness route is compiled into level geometry and is meant to match a designer-specified pacing, we discourage unnecessary detours (even if feasible) using explicit length and event-spacing constraints rather than relying on any suboptimal path. With our current pipeline reliably producing valid high-dimensional levels, we can also curate a large supervised dataset to explore PCGML and RL on top of HDPCG.
Besides, we will move from per-axis studies to axis composition, combining layers, time, locomotion modes, and other mechanisms in a single dimensional-expanded validator, so designers can specify coupled targets and obtain jointly consistent solutions. In parallel, we will include explicit robustness terms and multi-objective optimization into the GA, and report Pareto fronts to balance composite quality with re-planning success. 
Finally, we plan to package the end-to-end toolchain and conduct controlled user studies to connect our metrics to perceived pacing, challenge, and enjoyment.
Overall, we hope HDPCG can serve as a foundation for pushing level generation beyond geometry toward controllable, mechanism-rich, and designer-aligned worlds.


\clearpage

\bibliographystyle{ACM-Reference-Format}
\bibliography{references}

@String{Computing = "Computing" }

@String{Computer = "{IEEE} Computer" }

@String{Springer = "Springer-Verlag" }

@article{togelius2011search,
  title={Search-based procedural content generation: A taxonomy and survey},
  author={Togelius, Julian and Yannakakis, Georgios N and Stanley, Kenneth O and Browne, Cameron},
  journal={IEEE Transactions on Computational Intelligence and AI in Games},
  volume={3},
  number={3},
  pages={172--186},
  year={2011},
  publisher={IEEE}
}

@article{summerville2018procedural,
  title={Procedural content generation via machine learning (PCGML)},
  author={Summerville, Adam and Snodgrass, Sam and Guzdial, Matthew and Holmg{\aa}rd, Christoffer and Hoover, Amy K and Isaksen, Aaron and Nealen, Andy and Togelius, Julian},
  journal={IEEE Transactions on Games},
  volume={10},
  number={3},
  pages={257--270},
  year={2018},
  publisher={IEEE}
}

@inproceedings{parish2001procedural,
  title={Procedural modeling of cities},
  author={Parish, Yoav IH and M{\"u}ller, Pascal},
  booktitle={Proceedings of the 28th annual conference on Computer graphics and interactive techniques},
  pages={301--308},
  year={2001}
}

@article{wonka2003instant,
  title={Instant architecture},
  author={Wonka, Peter and Wimmer, Michael and Sillion, Fran{\c{c}}ois and Ribarsky, William},
  journal={ACM Transactions on Graphics (TOG)},
  volume={22},
  number={3},
  pages={669--677},
  year={2003},
  publisher={ACM New York, NY, USA}
}

@article{dormans2011generating,
  title={Generating missions and spaces for adaptable play experiences},
  author={Dormans, Joris and Bakkes, Sander},
  journal={IEEE Transactions on Computational Intelligence and AI in Games},
  volume={3},
  number={3},
  pages={216--228},
  year={2011},
  publisher={IEEE}
}

@inproceedings{smith2010tanagra,
  title={Tanagra: A mixed-initiative level design tool},
  author={Smith, Gillian and Whitehead, Jim and Mateas, Michael},
  booktitle={Proceedings of the Fifth International Conference on the Foundations of Digital Games},
  pages={209--216},
  year={2010}
}

@article{smith2010launchpad,
  title={Launchpad: A rhythm-based level generator for 2-d platformers},
  author={Smith, Gillian and Whitehead, Jim and Mateas, Michael and Treanor, Mike and March, Jameka and Cha, Mee},
  journal={IEEE Transactions on computational intelligence and AI in games},
  volume={3},
  number={1},
  pages={1--16},
  year={2010},
  publisher={IEEE}
}

@inproceedings{smith2009rhythm,
  title={Rhythm-based level generation for 2D platformers},
  author={Smith, Gillian and Treanor, Mike and Whitehead, Jim and Mateas, Michael},
  booktitle={Proceedings of the 4th international Conference on Foundations of Digital Games},
  pages={175--182},
  year={2009}
}

@inproceedings{jiang2022learning,
  title={Learning controllable 3D level generators},
  author={Jiang, Zehua and Earle, Sam and Green, Michael and Togelius, Julian},
  booktitle={Proceedings of the 17th International Conference on the Foundations of Digital Games},
  pages={1--9},
  year={2022}
}

@inproceedings{facey2024toward,
  title={Toward space-time WaveFunctionCollapse for level and solution generation},
  author={Facey, Kaylah and Cooper, Seth},
  booktitle={Proceedings of the AAAI Conference on Artificial Intelligence and Interactive Digital Entertainment},
  volume={20},
  number={1},
  pages={25--34},
  year={2024}
}

@inproceedings{Vandara2025SpacetimeLevelGenerationEditing,
  author    = {Akshar Vandara and Kaylah Facey and Seth Cooper},
  title     = {Spacetime Level Generation and Editing with Constraints from Examples},
  booktitle = {Proceedings of the AAAI Conference on Artificial Intelligence and Interactive Digital Entertainment (AIIDE)},
  year      = {2025},
  month     = nov,
}

@article{hoppe2000quickest,
  title={The quickest transshipment problem},
  author={Hoppe, Bruce and Tardos, {\'E}va},
  journal={Mathematics of Operations Research},
  volume={25},
  number={1},
  pages={36--62},
  year={2000},
  publisher={INFORMS}
}

@inproceedings{kohler2002time,
  title={Time-expanded graphs for flow-dependent transit times},
  author={K{\"o}hler, Ekkehard and Langkau, Katharina and Skutella, Martin},
  booktitle={European symposium on algorithms},
  pages={599--611},
  year={2002},
  organization={Springer}
}

@article{wang2019time,
  title={Time-dependent graphs: Definitions, applications, and algorithms},
  author={Wang, Yishu and Yuan, Ye and Ma, Yuliang and Wang, Guoren},
  journal={Data Science and Engineering},
  volume={4},
  number={4},
  pages={352--366},
  year={2019},
  publisher={Springer}
}

@article{erdmann1987multiple,
  title={On multiple moving objects},
  author={Erdmann, Michael and Lozano-Perez, Tomas},
  journal={Algorithmica},
  volume={2},
  number={1},
  pages={477--521},
  year={1987},
  publisher={Springer}
}

@inproceedings{shaker2013evolving,
  title={Evolving playable content for cut the rope through a simulation-based approach},
  author={Shaker, Noor and Shaker, Mohammad and Togelius, Julian},
  booktitle={Proceedings of the AAAI Conference on Artificial Intelligence and Interactive Digital Entertainment},
  volume={9},
  number={1},
  pages={72--78},
  year={2013}
}

@inproceedings{stephenson2016procedural,
  title={Procedural generation of levels for angry birds style physics games},
  author={Stephenson, Matthew and Renz, Jochen},
  booktitle={Proceedings of the AAAI Conference on Artificial Intelligence and Interactive Digital Entertainment},
  volume={12},
  number={1},
  pages={225--231},
  year={2016}
}

@inproceedings{ferreira2014generating,
  title={Generating levels for physics-based puzzle games with estimation of distribution algorithms},
  author={Ferreira, Lucas and Toledo, Claudio},
  booktitle={Proceedings of the 11th Conference on Advances in Computer Entertainment Technology},
  pages={1--6},
  year={2014}
}

@inproceedings{xu2014generative,
  title={Generative methods for guard and camera placement in stealth games},
  author={Xu, Qihan and Tremblay, Jonathan and Verbrugge, Clark},
  booktitle={Proceedings of the AAAI Conference on Artificial Intelligence and Interactive Digital Entertainment},
  volume={10},
  number={1},
  pages={87--93},
  year={2014}
}

@inproceedings{earle2025dreamgarden,
author = {Earle, Sam and Parajuli, Samyak and Banburski-Fahey, Andrzej},
title = {DreamGarden: A Designer Assistant for Growing Games from a Single Prompt},
year = {2025},
isbn = {9798400713941},
publisher = {Association for Computing Machinery},
address = {New York, NY, USA},
url = {https://doi.org/10.1145/3706598.3714233},
doi = {10.1145/3706598.3714233},
booktitle = {Proceedings of the 2025 CHI Conference on Human Factors in Computing Systems},
articleno = {57},
numpages = {19},
series = {CHI '25}
}

@inproceedings{hu20243d,
  title={3d building generation in minecraft via large language models},
  author={Hu, Shiying and Huang, Zengrong and Hu, Chengpeng and Liu, Jialin},
  booktitle={2024 IEEE Conference on Games (CoG)},
  pages={1--4},
  year={2024},
  organization={IEEE}
}

@article{kobenova2024social,
  title={Social Conjuring: Multi-User Runtime Collaboration with AI in Building Virtual 3D Worlds},
  author={Kobenova, Amina and DeVeaux, Cyan and Parajuli, Samyak and Banburski-Fahey, Andrzej and Fernandez, Judith Amores and Lanier, Jaron},
  journal={arXiv preprint arXiv:2410.00274},
  year={2024}
}

@inproceedings{earle2024dreamcraft,
  title={Dreamcraft: Text-guided generation of functional 3D environments in Minecraft},
  author={Earle, Sam and Kokkinos, Filippos and Nie, Yuhe and Togelius, Julian and Raileanu, Roberta},
  booktitle={Proceedings of the 19th International Conference on the Foundations of Digital Games},
  pages={1--15},
  year={2024}
}

@inproceedings{xu2025constraint,
author = {Xu, Kaijie and Verbrugge, Clark},
title = {Constraint Is All You Need: Optimization-Based 3D Level Generation with LLMs},
year = {2025},
isbn = {9798400718564},
publisher = {Association for Computing Machinery},
address = {New York, NY, USA},
url = {https://doi.org/10.1145/3723498.3723840},
doi = {10.1145/3723498.3723840},
articleno = {66},
numpages = {13},
series = {FDG '25}
}

@article{xu2025database,
  title={A Database-Driven Framework for 3D Level Generation with LLMs},
  author={Xu, Kaijie and Verbrugge, Clark},
  journal={arXiv preprint arXiv:2508.18533},
  year={2025}
}

@INPROCEEDINGS{xu2025actionwindow,
  author={Xu, Kaijie and Verbrugge, Clark},
  booktitle={2025 IEEE Conference on Games (CoG)}, 
  title={Action Window Planning for Stealth Missions}, 
  year={2025},
  volume={},
  number={},
  pages={1-4},
  doi={10.1109/CoG64752.2025.11114360}}

@article{shaker2016procedural,
  title={Procedural content generation in games},
  author={Shaker, Noor and Togelius, Julian and Nelson, Mark J},
  year={2016},
  publisher={Springer}
}

@article{yannakakis2011experience,
  title={Experience-driven procedural content generation},
  author={Yannakakis, Georgios N and Togelius, Julian},
  journal={IEEE Transactions on Affective Computing},
  volume={2},
  number={3},
  pages={147--161},
  year={2011},
  publisher={IEEE}
}

@article{khatib1986real,
  title={Real-time obstacle avoidance for manipulators and mobile robots},
  author={Khatib, Oussama},
  journal={The international journal of robotics research},
  volume={5},
  number={1},
  pages={90--98},
  year={1986},
  publisher={Sage Publications Sage CA: Thousand Oaks, CA}
}

@article{rimon1992exact,
  title={Exact robot navigation using artificial potential functions},
  author={Rimon, E and Koditschek, DE},
  journal={IEEE Transactions on Robotics and Automation},
  volume={8},
  number={5},
  pages={501--518},
  year={1992},
  publisher={IEEE}
}

@article{eppstein1998finding,
  title={Finding the k shortest paths},
  author={Eppstein, David},
  journal={SIAM Journal on computing},
  volume={28},
  number={2},
  pages={652--673},
  year={1998},
  publisher={SIAM}
}

@article{cliff1993dominance,
  title={Dominance statistics: Ordinal analyses to answer ordinal questions.},
  author={Cliff, Norman},
  journal={Psychological bulletin},
  volume={114},
  number={3},
  pages={494},
  year={1993},
  publisher={American Psychological Association}
}

@book{cliff2014ordinal,
  title={Ordinal methods for behavioral data analysis},
  author={Cliff, Norman},
  year={2014},
  publisher={Psychology Press}
}

@article{hart1968formal,
  title={A formal basis for the heuristic determination of minimum cost paths},
  author={Hart, Peter E and Nilsson, Nils J and Raphael, Bertram},
  journal={IEEE transactions on Systems Science and Cybernetics},
  volume={4},
  number={2},
  pages={100--107},
  year={1968},
  publisher={IEEE}
}

@inproceedings{van2011reciprocal,
  title={Reciprocal n-body collision avoidance},
  author={Van Den Berg, Jur and Guy, Stephen J and Lin, Ming and Manocha, Dinesh},
  booktitle={Robotics Research: The 14th International Symposium ISRR},
  pages={3--19},
  year={2011},
  organization={Springer}
}

@book{holland1992adaptation,
  title={Adaptation in natural and artificial systems: an introductory analysis with applications to biology, control, and artificial intelligence},
  author={Holland, John H},
  year={1992},
  publisher={MIT press}
}

@inproceedings{dormans2011level,
  title={Level design as model transformation: a strategy for automated content generation},
  author={Dormans, Joris},
  booktitle={Proceedings of the 2nd International Workshop on Procedural Content Generation in Games},
  pages={1--8},
  year={2011}
}

@inproceedings{fernandes2024generating,
  title={Generating game levels by defining player experiences},
  author={Fernandes, Pedro M and Lopes, Manuel and Prada, Rui},
  booktitle={Proceedings of the AAAI Conference on Artificial Intelligence and Interactive Digital Entertainment},
  volume={20},
  number={1},
  pages={179--188},
  year={2024}
}

@inproceedings{van2015procedural,
  title={Procedural generation of collaborative puzzle-platform game levels},
  author={van Arkel, Benjamin and Karavolos, Daniel and Bouwer, Anders},
  booktitle={GAME ON'2015: 16th International Conference on Intelligent Games and Simulation},
  pages={87--93},
  year={2015},
  organization={Eurosis}
}

@article{zagal2010time,
  title={Time in video games: A survey and analysis},
  author={Zagal, Jos{\'e} P and Mateas, Michael},
  journal={Simulation \& Gaming},
  volume={41},
  number={6},
  pages={844--868},
  year={2010},
  publisher={SAGE Publications Sage CA: Los Angeles, CA}
}

@inproceedings{van2011navigation,
  title={Navigation meshes for realistic multi-layered environments},
  author={Van Toll, Wouter and Cook, Atlas F and Geraerts, Roland},
  booktitle={2011 IEEE/RSJ International Conference on Intelligent Robots and Systems},
  pages={3526--3532},
  year={2011},
  organization={IEEE}
}

@article{al2018virtual,
  title={Virtual locomotion: a survey},
  author={Al Zayer, Majed and MacNeilage, Paul and Folmer, Eelke},
  journal={IEEE transactions on visualization and computer graphics},
  volume={26},
  number={6},
  pages={2315--2334},
  year={2018},
  publisher={IEEE}
}

@article{anderton2025teleportation,
  title={From teleportation to climbing: A review of locomotion techniques in the most used commercial virtual reality applications},
  author={Anderton, Craig and Creed, Chris and Sarcar, Sayan and Theil, Arthur},
  journal={International Journal of Human--Computer Interaction},
  volume={41},
  number={4},
  pages={1946--1966},
  year={2025},
  publisher={Taylor \& Francis}
}

@inproceedings{hans2020spaces,
  title={Spaces of Allegory. Non-Euclidean Spatiality as a Ludo-Poetic Device},
  author={HANS-JOACHIM, BACKE},
  booktitle={DiGRA-Proceedings of the 2020 DiGRA International Conference: Play Everywhere},
  volume={2},
  year={2020}
}

@inproceedings{karth2017wavefunctioncollapse,
  title={WaveFunctionCollapse is constraint solving in the wild},
  author={Karth, Isaac and Smith, Adam M},
  booktitle={Proceedings of the 12th International Conference on the Foundations of Digital Games},
  pages={1--10},
  year={2017}
}

@article{karth2021wavefunctioncollapse,
  title={Wavefunctioncollapse: Content generation via constraint solving and machine learning},
  author={Karth, Isaac and Smith, Adam M},
  journal={IEEE Transactions on Games},
  volume={14},
  number={3},
  pages={364--376},
  year={2021},
  publisher={IEEE}
}

@inproceedings{cooper2023sturgeon,
  title={Sturgeon-MKIII: Simultaneous level and example playthrough generation via constraint satisfaction with tile rewrite rules},
  author={Cooper, Seth},
  booktitle={Proceedings of the 18th International Conference on the Foundations of Digital Games},
  pages={1--9},
  year={2023}
}

@inproceedings{cooper2024sturgeon,
  title={Sturgeon-MKIV: constraint-based level and playthrough generation with graph label rewrite rules},
  author={Cooper, Seth and Bazzaz, Mahsa},
  booktitle={Proceedings of the AAAI Conference on Artificial Intelligence and Interactive Digital Entertainment},
  volume={20},
  number={1},
  pages={13--24},
  year={2024}
}

@incollection{kovar2023motion,
  title={Motion graphs},
  author={Kovar, Lucas and Gleicher, Michael and Pighin, Fr{\'e}d{\'e}ric},
  booktitle={Seminal Graphics Papers: Pushing the Boundaries, Volume 2},
  pages={723--732},
  year={2023}
}

@article{liu2005learning,
  title={Learning physics-based motion style with nonlinear inverse optimization},
  author={Liu, C Karen and Hertzmann, Aaron and Popovi{\'c}, Zoran},
  journal={ACM Transactions on Graphics (TOG)},
  volume={24},
  number={3},
  pages={1071--1081},
  year={2005},
  publisher={ACM New York, NY, USA}
}

@article{witkin1988spacetime,
  title={Spacetime constraints},
  author={Witkin, Andrew and Kass, Michael},
  journal={ACM Siggraph Computer Graphics},
  volume={22},
  number={4},
  pages={159--168},
  year={1988},
  publisher={ACM New York, NY, USA}
}

@inproceedings{cooper2025constraint,
  author    = {Seth Cooper and Mahsa Bazzaz},
  title     = {A Constraint-Based Graph Grammar Approach Unifying Level and Playthrough Generation},
  booktitle = {EXAG-INT 2025: Experimental AI in Games and Intelligent Narrative Technologies 2025},
  series    = {CEUR Workshop Proceedings},
  volume    = {4090},
  address   = {Edmonton, Alberta, Canada},
  year      = {2025}
}

\appendix
\section{TEG-A* Details}
\label{app:astar}

\subsection{Exact TEG-A*}
Let $\mathcal{X}\subset\mathbb{Z}^2$ and time domain be $\mathcal{T}=\{0,\dots,T_{\max}\}$ or $\mathbb{Z}_T$. A state augments space-time with interaction memory:
$
s=(x,t, z), z\in\mathcal{Z},
$
where $z$ records which platforms have been used and which obstacle endpoints have been hit (bitmasks or small counters). Edges $e=(s\!\to\!s')$ implement \emph{wait}, \emph{walk}, and \emph{ride} under feasibility $F$ (endpoint-only boarding/alighting, occupancy and head-on swap checks). To \emph{drive} non-trivial yet non-redundant interactions, the edge cost
$
c(e)=c_{\text{time}}(e)-\lambda_{\text{ride}}\cdot\mathbf{1}_{\text{ride}}(e)-\lambda_{\text{hit}}\cdot\Delta\text{hit}(z\!\to\!z'),
$
rewards riding and the \emph{first} hits on required endpoints; a goal test demands $x=x_g$ and $z\in\mathcal{Z}_{\text{accept}}$ (e.g., quotas or full coverage). The admissible heuristic $h(x,t)\!=\!\|x-x_g\|_1$ lower-bounds remaining time.

\textbf{Complexity:} Let $|\mathcal{X}|=W\!\times\!H$, time size $|\mathcal{T}|=T$, number of platforms $P$, and required obstacle endpoints $E$. Even with bitmasks, the state space scales as
$
|\widetilde{\mathcal{S}}|=\mathcal{O}\!\big(WHT\cdot 2^{P}\cdot 2^{E}\big),
$
and branching $b\approx 5 + \Theta(P)$ (four moves + wait + ride options), yielding a worst-case expansion of $\Theta(b\cdot|\widetilde{\mathcal{S}}|)$. Moreover, waiting typically expands step-by-step, making practical node counts proportional to long dwell times near boarding windows. In our setting, this leads to timeouts even on small grids when interaction shaping is enabled.

\subsection{Simplified TEG-A*}
Pseudo-code is shown in Algorithm~\ref{alg:astar}. To remain tractable while preserving feasibility semantics, we adopt a cyclic TEG with lightweight memory and coarse costs. The state is
$
s=(t\bmod T, x, y, \allowbreak pm, ep),
$
where pm and ep are bitmasks for ``used platforms'' and ``visited obstacle endpoints.'' Time wraps modulo the LCM $T$ of object cycles. We apply:

\begin{enumerate}
\item \textbf{Feasibility rules (unchanged).} Endpoint-only riding; per-step occupancy and swap checks; no mid-track walking (non-endpoint track cells are non-walkable).
\item \textbf{Coarse costs and heuristic.} Unit cost for walk/wait; ride cost equals ride duration; $h(x,t)=\|x-x_g\|_1$.
\item \textbf{Lightweight search policy.} Single-close (no re-opening states) to cap memory and runtime; cyclic time avoids large $T_{\max}$.
\item \textbf{Goal (hard) condition.} Reach $x_g$ with all platforms used and all sampled obstacle endpoints visited (pm and ep match their full masks).
\end{enumerate}

\begin{algorithm}[t]
\caption{Simplified TEG-A*}
\begin{algorithmic}[1]
\State Build periodic objects; let $T=\mathrm{lcm}(\text{cycles})$.
\State Start $s_0=(0,x_s,y_s,0,ep_0)$; push to PQ with $f=g+h$.
\While{PQ not empty}
  \State Pop $(f,g,s=(t,x,y,pm,ep))$
  \If{$(x,y)=x_g$ \textbf{and} $pm=ALL\_P$ \textbf{and} $ep=ALL\_EP$} \Return path \EndIf
  \For{each move/wait $(dx,dy)\in\{4\text{-neigh},(0,0)\}$}
     \State $s'\leftarrow (t{+}1\bmod T, x{+}dx, y{+}dy, pm, ep')$
     \State \textbf{continue} if out of bounds, cliff, mid-track, occupied at $t{+}1$, or swap
     \State push $s'$ with $g{+}1, h=\|x'-x_g\|_1$
  \EndFor
  \For{each platform $i$ with endpoint at $(x,y)$ and $p_i(t)=(x,y)$}
     \State Let $t'$ be half- or full-cycle to the opposite endpoint; \textbf{continue} if any frame collides
     \State $s'\!\leftarrow (t'{\bmod}T, x', y', pm\,|\,2^i, ep')$; push with cost $g{+}(t'-t)$
  \EndFor
\EndWhile
\State \Return infeasible
\end{algorithmic}
\label{alg:astar}
\end{algorithm}

\textbf{Rationality and correctness:} If the algorithm returns a path, per-step occupancy and swap checks plus endpoint-only riding ensure the trajectory satisfies $F$; the reconstruction enforces no mid-track walking and avoids spatial self-collisions. The single-close policy sacrifices optimality but preserves \emph{feasibility}. Coarse costs and cyclic time make the search compact; however, without explicit temporal rewards/quotas, the solver does not reliably \emph{induce} interactions—exactly the empirical behavior we report. These limitations motivate our use of \textbf{TEG-DP} for high-quality, controllable time-aware generation at scale.

\section{DP Optimality on the Time-Expanded Graph}
\label{app:dp}

Let $X\subset\mathbb{Z}^2$ be the spatial lattice and $T=\{0,\dots,\tau\}$ a finite horizon. The time-expanded graph is the layered DAG $\widetilde{\mathcal{G}}=(\mathcal{V},\mathcal{E})$ with layers $\mathcal{V}_t=\{(x,t):x\in X\ \wedge\ (x,t)\ \text{allowed}\}$ and edges $\mathcal{E}_t\subseteq\mathcal{V}_t\times\mathcal{V}_{t+1}$ encoding \textsc{Wait}, four-connected \textsc{Walk}, and endpoint-only \textsc{Ride}. Each edge $e\in\mathcal{E}_t$ has nonnegative additive cost $c_t(e)\ge 0$; forbidden states are removed (or assigned $+\infty$). The start set is $S_0\subseteq\mathcal{V}_0$ and the goal set is $G_\tau\subseteq\mathcal{V}_\tau$.

Define $V_t(x)$ as the minimum cost to reach $(x,t)$ from any $(x_0,0)\in S_0$, with initialization $V_0(x)=0$ for $(x,0)\in S_0$ and $V_0(x)=+\infty$ otherwise. The forward recurrence is
\[
V_{t+1}(x')=\min_{(x\to x')\in\mathcal{E}_t}\big\{V_t(x)+c_t(x\to x')\big\},\qquad t=0,\dots,\tau-1,
\]
and a backpointer is stored at each update to reconstruct a terminal path in $G_\tau$. Because $\widetilde{\mathcal{G}}$ is acyclic and the objective is additive, optimal substructure holds: if a path to $(x',t{+}1)$ is globally optimal, its prefix to any $(x,t)$ on that path is also optimal. By induction over layers, the recurrence computes the exact minimum $V_t(\cdot)$ on every layer; selecting $\min_{(x_\mathrm{g},\tau)\in G_\tau}V_\tau(x_\mathrm{g})$ yields a globally minimum-cost path. Runtime is $\mathcal{O}(|\mathcal{E}|)$ and memory is $\mathcal{O}(|\mathcal{V}|)$.

Setting $c_t(e)\equiv 1$ reduces the recurrence to unweighted shortest path on the layered TEG; any layer-wise breadth-first search returns an equal-length optimum (ties aside). Using admissible and consistent heuristics, A* on the same TEG returns the same optimum; simplified A* variants that employ coarse costs or single-close policies need not be optimal, whereas dynamic programming on the layered DAG remains exact.

Periodic dynamics are handled by unrolling a finite horizon $\tau$ (e.g., a fixed $T_{\max}$ per scale or the least common multiple of object periods), which preserves acyclicity; optimality above is therefore with respect to the chosen horizon. Optimality also presumes Markovian feasibility and costs in $(x,t)$; history-dependent constraints would require state augmentation to retain optimal substructure, which we do not use in our validators.

\section{Reproducibility}
\label{app:repro}

Experiments are deterministic per run by fixing seed = run\_index. For Space: single mode uses 80/40/20 seeds at S/M/L, GA mode uses 40/20/10 seeds at S/M/L. For Time: both single and GA use 12/8/4 seeds at S/M/L. Approximate memory needs are $\sim$4\,GB for Space-L ($100^3$ with two layers) and $\sim$2\,GB for Time-L ($80\times 40$ with $T_{\max}=500$).

Space validation uses a 6-connected (Von Neumann) neighborhood and an L1 heuristic. NNB adds an i.i.d.\ penalty field drawn from $\mathrm{Uniform}(0,50)$ and uses the raw shortest-path skeleton; $K$ switches are sampled at near-uniform indices excluding endpoints. NP–A* adds linear-decay repulsive blobs of radius~4 with peak penalty $10\,000$, plus scale-specific cost blobs: S: 10 blobs with radius $[1.5,2.5]$ and cost $[20,50]$; M: 15 blobs, $[2.0,3.0]$, $[30,60]$; L: 25 blobs, $[3.0,4.0]$, $[40,80]$. PF–A* combines repulsive blobs with attractive switch anchors by subtracting a switch reward of 200 at S/M and 300 at L; if a cross-layer “teleport” is detected, a local in-layer patch is inserted before switching. Postplanning growth enforces component-aware corridors then rooms, with conflict checks across in-layer neighbors, same-cell/same-layer, in-layer adjacency during candidate expansion, and cross-layer coincidences; switch steps open the target layer, locally close the source layer, and keep the switch cell bi-layer open. 4D verification runs A* on $(x,y,z,\ell)$ with $h=\lVert x-x_g\rVert_1$ (admissible with nonnegative switch costs), unit spatial move cost, and layer-switch cost $\in\{1,2\}$. Controllability sweeps use S: $\rho\in\{1,3,5\}$ and $s_{\min}\in\{3,5,7\}$; M: $\rho\in\{2,4\}$ and $s_{\min}\in\{4,6\}$; L: $\rho=2.0$, $s_{\min}=5$. Start–goal endpoints are sampled with minimum Manhattan distances 20/25/30 at S/M/L, up to 1000 attempts, with 12/10/12 accepted pairs per target at S/M/L. Robustness perturbations use a band protocol at S/M (radius $r=1$, closing open voxels with probability $p=0.01$ along the nominal path) and a global protocol at L ($p=0.005$ over the full grid), with fixed repeats per target/method/scale. The Space GA uses population~15, generations 30/20/10 at S/M/L, mutation rate~0.3, and tournament size~3; chromosomes encode the Stage–1 parameters (blob counts/radii/weights, switch anchors/indices, skeleton seed), and fitness is the composite spatial score.

Time validation uses a time-expanded graph with $\Delta t=1$ and finite horizons $T_{\max}=200/300/500$ at S/M/L. Actions are \textsc{Wait} $(x,y,t)\!\to\!(x,y,t{+}1)$, four-connected \textsc{Walk} $(x,y,t)\!\to\!(x',y',t{+}1)$, and endpoint-only \textsc{Ride} along platform trajectories; per-tick occupancy is enforced and swap-prevention forbids cross-occupancy. Static scenarios instantiate 3/5/8 platforms and 3/5/8 obstacles at S/M/L with zig-line back-and-forth trajectories, minimum lengths 4/5/6 for platforms and 4–6 for obstacles, unit speed, and period $2L{-}2$. TEG–A* shares catalogs, restricts boarding to endpoints (start and midpoint ticks), requires at least three ride ticks, and uses coarse costs with a single-close policy; paths are feasible but not guaranteed optimal. TEG–DP builds per-time forbidden masks and a nonnegative cost field $c_t$ and solves the layered DAG by forward min-cost dynamic programming
\[
V_{t+1}(x')=\min_{(x\to x')\in E_t,\ \neg\mathrm{forbid}_{t+1}(x')}\{V_t(x)+c_t(x\!\to\!x')\},
\]\[
V_0(x)=0\ \text{on starts},\ +\infty\ \text{otherwise},
\]
returning the global minimum-cost path on the finite-horizon TEG; $c_t\equiv 1$ yields the unweighted shortest path (layer-wise BFS). 
\begingroup\abovedisplayskip=4pt\belowdisplayskip=4pt
\begin{align}
c_t(x\!\to\!x') \;=&\; 
c_{\textsc{Walk}}\mathbb{1}_{\textsc{Walk}}
+c_{\textsc{Wait}}\mathbb{1}_{\textsc{Wait}}
+c_{\textsc{Ride}}\mathbb{1}_{\textsc{Ride}} \nonumber\\
&\;+\;\lambda_{\mathrm{cad}}\;\mathbb{1}_{\textsc{Ride}}\;\mathbb{1}\{t\notin \mathcal{W}_{r^\star}(P)\}
+\lambda_{\mathrm{uni}}\;\big|\,s_t-\mathbb{1}_{\textsc{Ride}}\,\big| \nonumber\\
&\;+\;\lambda_{\mathrm{saf}}\;\mathrm{Danger}_t(x')\;
-\;\eta_{\mathrm{ep}}\;\mathbb{1}_{\text{Board/Alight}}\,,
\label{eq:ct}
\end{align}
\endgroup

\noindent
Constants used in all experiments: $(c_{\textsc{Walk}},c_{\textsc{Wait}},c_{\textsc{Ride}})=(1.0,1.0,\allowbreak0.25)$;\;
$(\lambda_{\mathrm{cad}},\lambda_{\mathrm{uni}},\lambda_{\mathrm{saf}},\eta_{\mathrm{ep}})=(0.5,0.3,2.0,0.1)$;\;
$P=d_{\min}+\{2,3,\allowbreak4\}_{\text{S/M/L}}$,\; $\omega=\mathrm{round}(r^\star P)$;\;
$s_t\in\{0,1\}$ shares $(P,\omega)$;\;
$\mathrm{Danger}_t(x')=1$ iff occupied or 4-neighbor at $t$, or would swap at $t{+}1$; else $0$.
All costs are clipped to be nonnegative before the DP update.
The Time GA uses population~20, generations~20, mutation rate~0.2, tournament size~3, and an interaction-guided fitness over ride ratio, spacing, beat regularity, interaction coverage and safety.

Space metrics include density/spacing controllability, alternative-route robustness, perturbation success rate and $\Delta$cost, and a composite spatial score as a weighted sum of five standardized terms with fixed weights $(0.5,1.0,2.0,2.0,1.5)$ for \{path length, length uniformity, spatial dispersion, minimal-length penalty, coverage\}; invalid runs set $\pm\infty$ and are filtered at aggregation. Time metrics include ride/wait ratios, total ticks, coverage, near-swap violations, transmission counts, and beat uniformity; ratios are reported to three decimals and time in integer ticks; undefined events return NaN and are ignored by aggregators. Statistical testing uses Mann–Whitney U with Cliff's $\delta$ and Bonferroni correction for Space \cite{cliff1993dominance, cliff2014ordinal}, and Mann–Whitney U with Cliff's $\delta$ and Holm–Bonferroni correction for Time; pairwise comparisons are conducted per experiment and run type, with CSV outputs for statistics, $p$-values, adjusted $p$-values, and effect sizes.
\section{Additional Quantitative Results for Space}
\label{app:space-more}

\begin{figure}[!t]
  \centering
  \includegraphics[width=\linewidth]{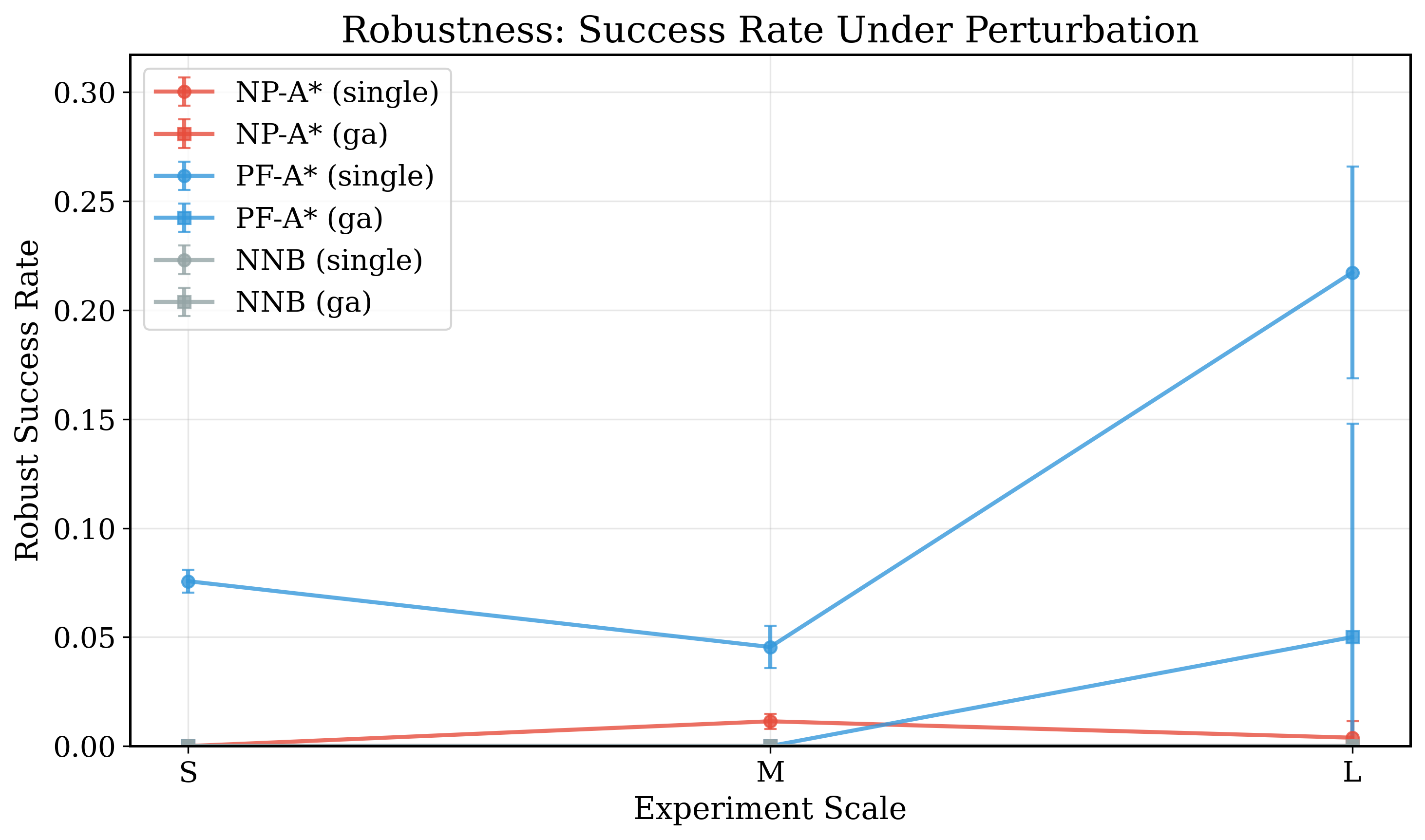}
  \caption{\textbf{Robustness to micro-perturbations.} PF-A* (single) attains the highest success on S/M/L; NP-A*/NNB are near zero. Error bars show across-seed std.}
  \label{fig:robust-rate}
\end{figure}

\begin{figure}[!t]
  \centering
  \includegraphics[width=\linewidth]{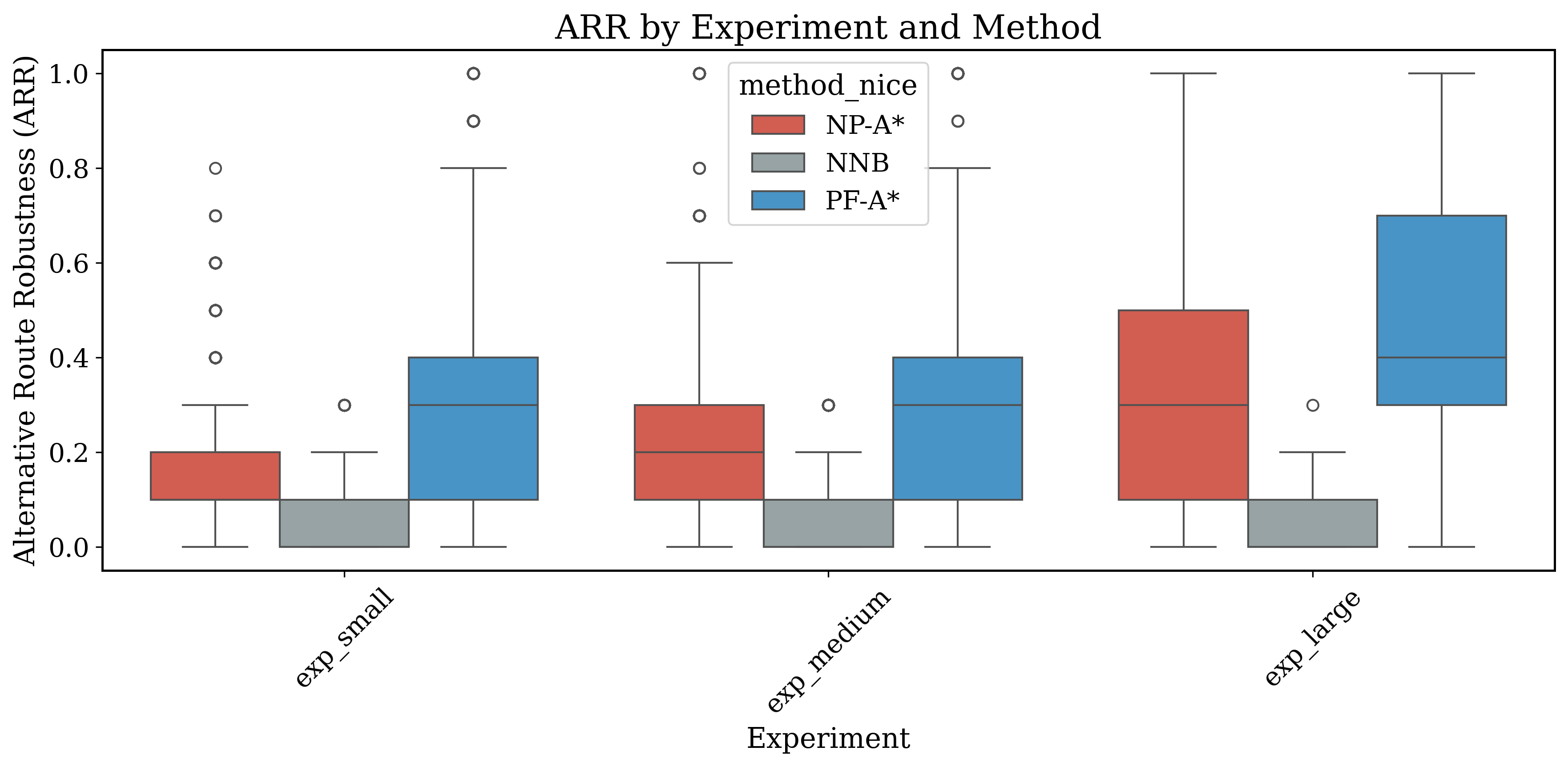}
  \caption{\textbf{ARR distributions (by method and scale).} PF-A* stochastically dominates NP-A* and NNB in single mode; GA narrows the gap on S/M and remains comparable on L.}
  \label{fig:arr-box}
\end{figure}

\begin{figure}[t]
  \centering
  \includegraphics[width=\linewidth]{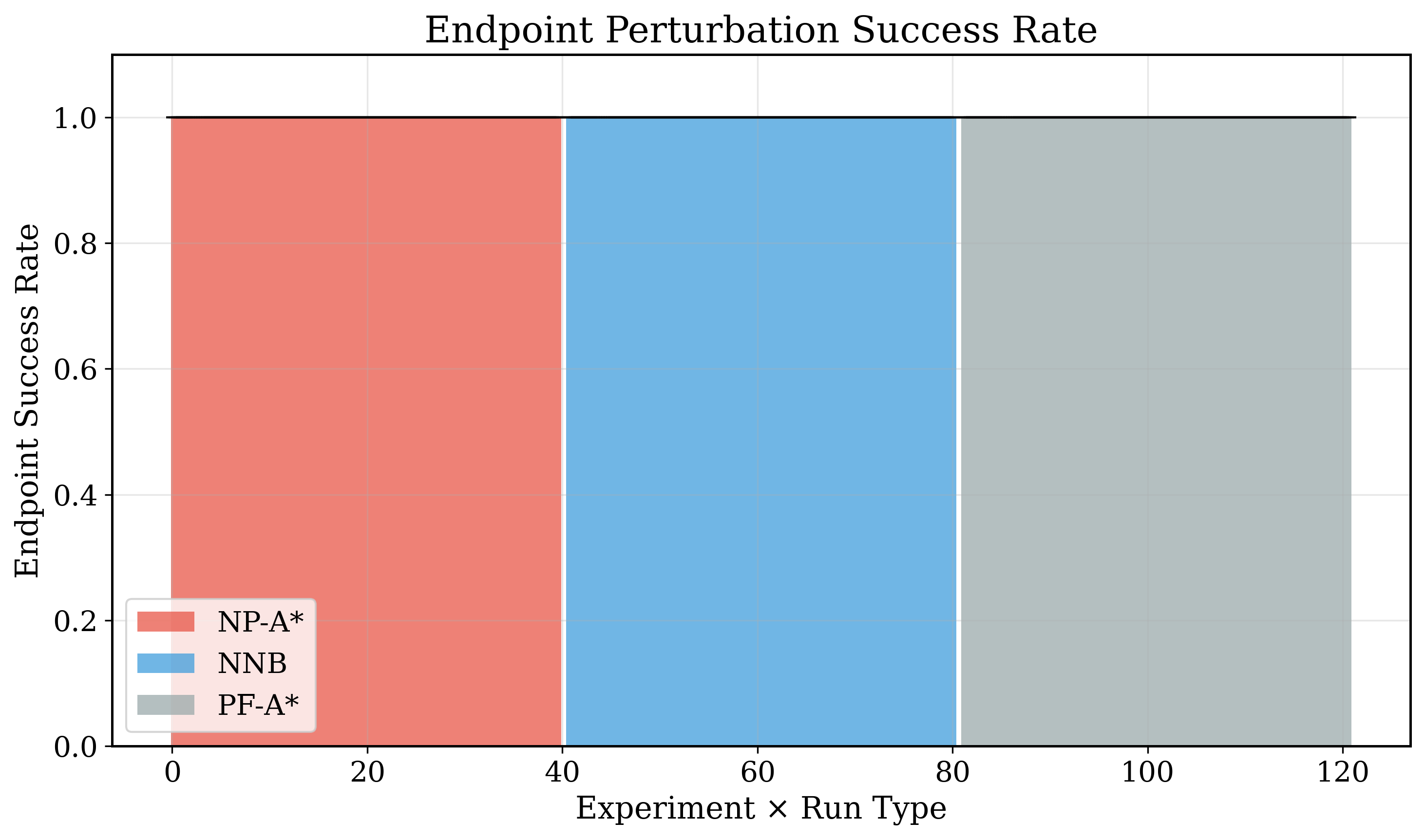}
  \caption{\textbf{Endpoint perturbations.} Near-ceiling success ($\approx 1.0$) for all methods and scales.}
  \label{fig:endpoint-success}
\end{figure}

\begin{figure}[t]
  \centering
  \includegraphics[width=\linewidth]{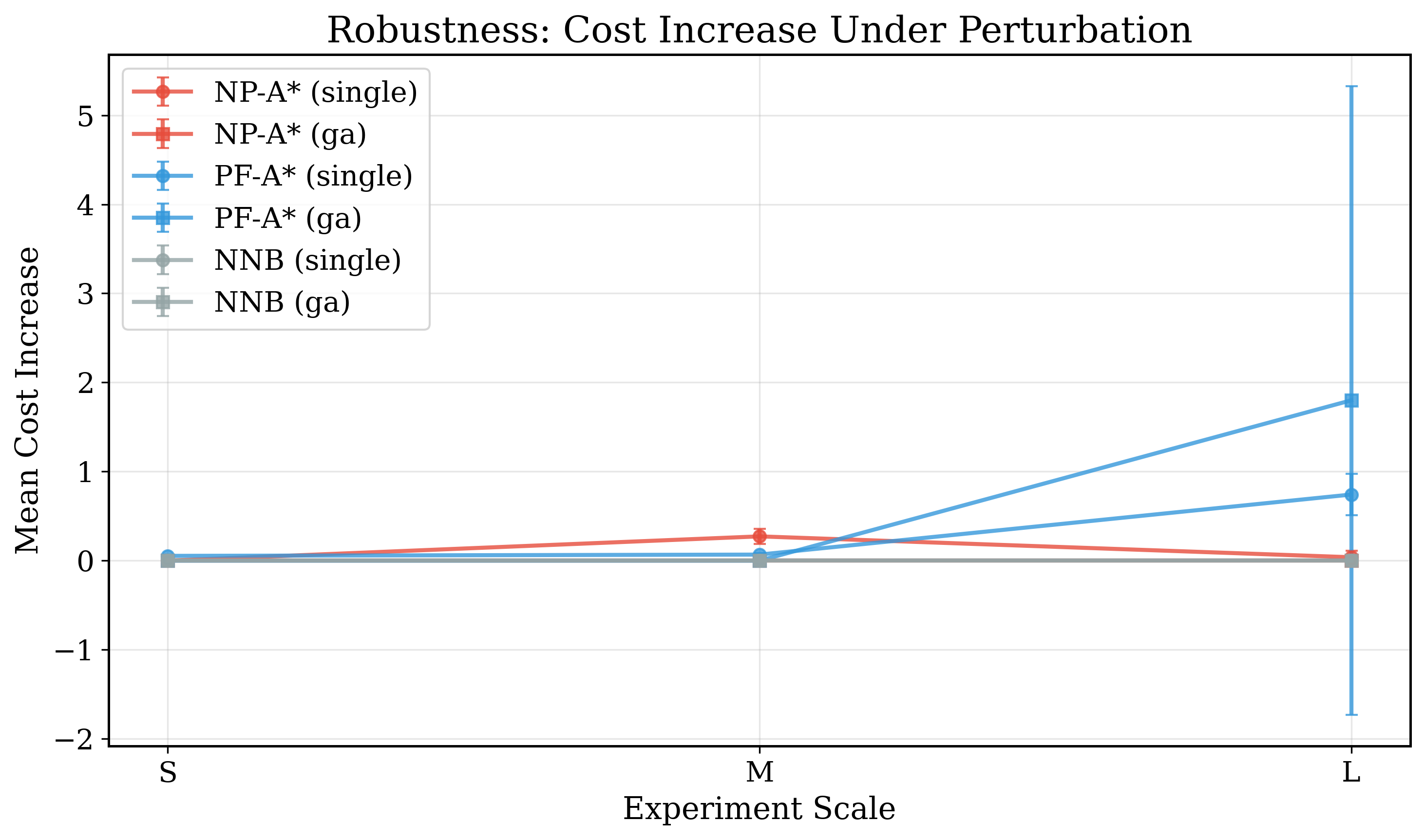}
  \caption{\textbf{Cost increase under micro-perturbations.} S/M: near-zero for all (except NP-A* at M); L: PF-A* accepts costlier detours while NP-A* stays conservative.}
  \label{fig:robust-cost}
\end{figure}

Beyond the scale-aggregated micro-perturbation success in Fig.~\ref{fig:robust-rate}, we provide distributional diagnostics and compact statistics that clarify the mechanisms behind R2--R3.

\textbf{Alternative Route Robustness (ARR).}
Fig.~\ref{fig:arr-box} shows ARR distributions; single-run means are:
\emph{PF-A*} $\{0.338,\,0.319,\,\mathbf{0.530}\}$ for S/M/L,
\emph{NP-A*} $\{0.156,\,0.215,\,0.360\}$,
\emph{NNB} $\{0.037,\,0.042,\,0.057\}$;
GA means are
PF-A* $\{0.138,\,0.245,\,0.320\}$,
NP-A* $\{0.233,\,\mathbf{0.370},\,0.340\}$,
NNB $\{0.075,\,0.070,\,0.050\}$.
Thus, \emph{single} consistently ranks PF-A* $>$ NP-A* $>$ NNB across scales, while \emph{GA} boosts NP-A* more on S/M and keeps both PF-A*/NP-A* comparable on L.

\textbf{Endpoint perturbations.}
Fig.~\ref{fig:endpoint-success} confirms near-ceiling success ($\approx 1.0$) for all methods and scales (each target set has $n{=}20$--$80$ replans), indicating that robustness failures mainly originate from local micro-structure changes rather than start/end moves.

\textbf{Perturbation success and cost trade.}
The micro-perturbation success rate (mean) for PF-A* (single) is
\{S: $\mathbf{0.0756}$, M: $\mathbf{0.0455}$, L: $\mathbf{0.2173}$\},
whereas NP-A* and NNB are near zero on all scales; GA variants are also near zero except PF-A* at L ($0.0500$).
Complementarily, Fig.~\ref{fig:robust-cost} summarize the mean $\Delta$cost (perturbed vs.\ nominal): on S/M all methods are close to $0$ except NP-A* at M ($0.273$); on L, PF-A* accepts noticeably costlier detours (single $0.742$, GA $1.800$) while NP-A* remains conservative ($0.038$) and NNB is $0$. 
Together with Fig.~\ref{fig:robust-rate}, this indicates that PF-A* achieves higher robustness partly by leveraging slightly more expensive alternatives on large grids, whereas NP-A* tends to reject them.

\begin{figure}[t]
  \centering
  \includegraphics[width=\linewidth]{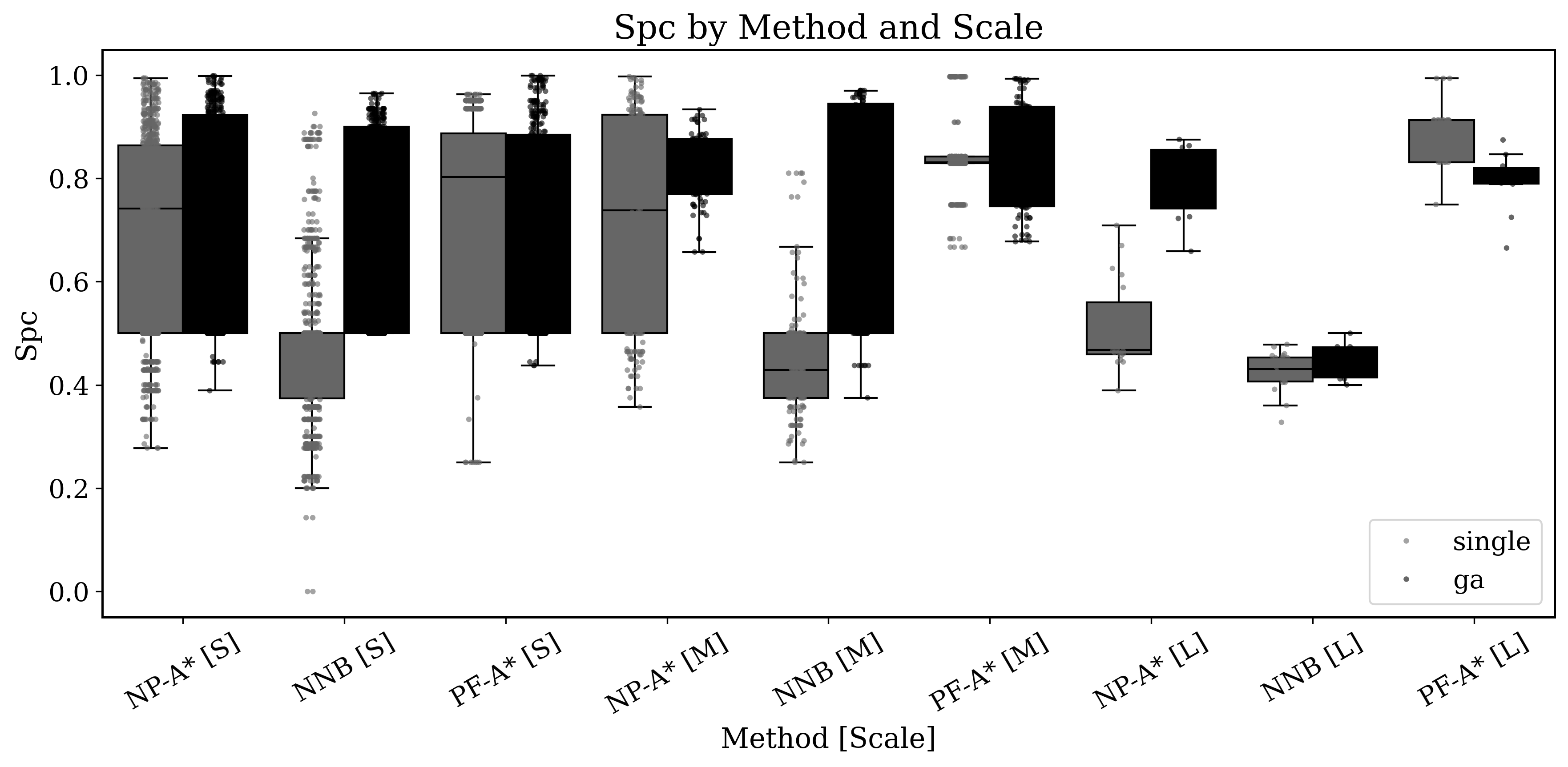}
  \caption{\textbf{Switch Pattern Compliance (SPC).} GA tightens patterns and ranks PF-A* highest on M; single-mode shows NP-A* higher on S/M and PF-A* catching up on L.}
  \label{fig:spc-box}
\end{figure}

\begin{figure}[t]
  \centering
  \includegraphics[width=\linewidth]{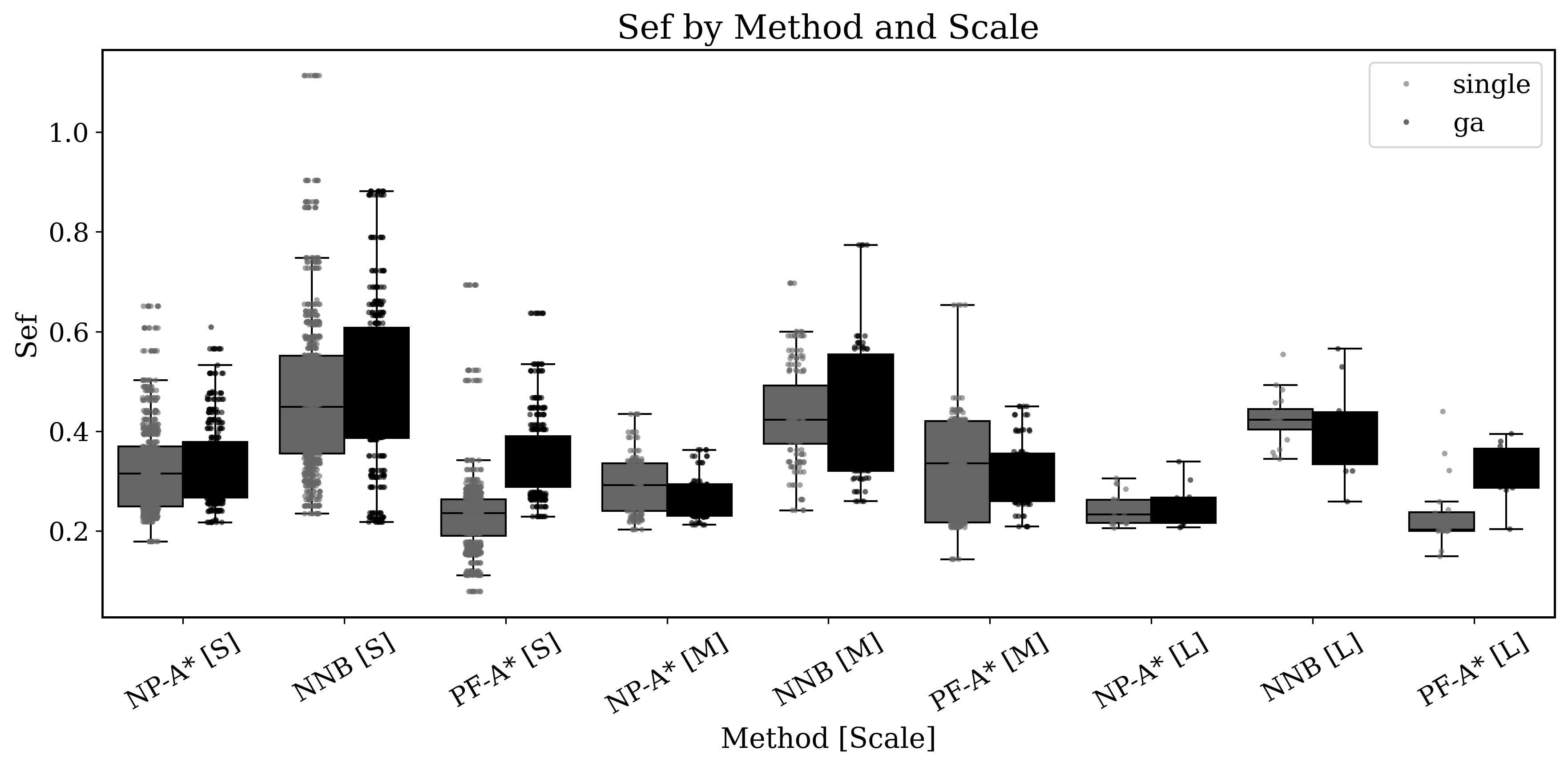}
  \caption{\textbf{Switch Exit Freedom (SEF).} NNB has higher local exit openness but without gains in controllability or robustness.}
  \label{fig:sef-box}
\end{figure}

\begin{figure}[t]
  \centering
  \includegraphics[width=\linewidth]{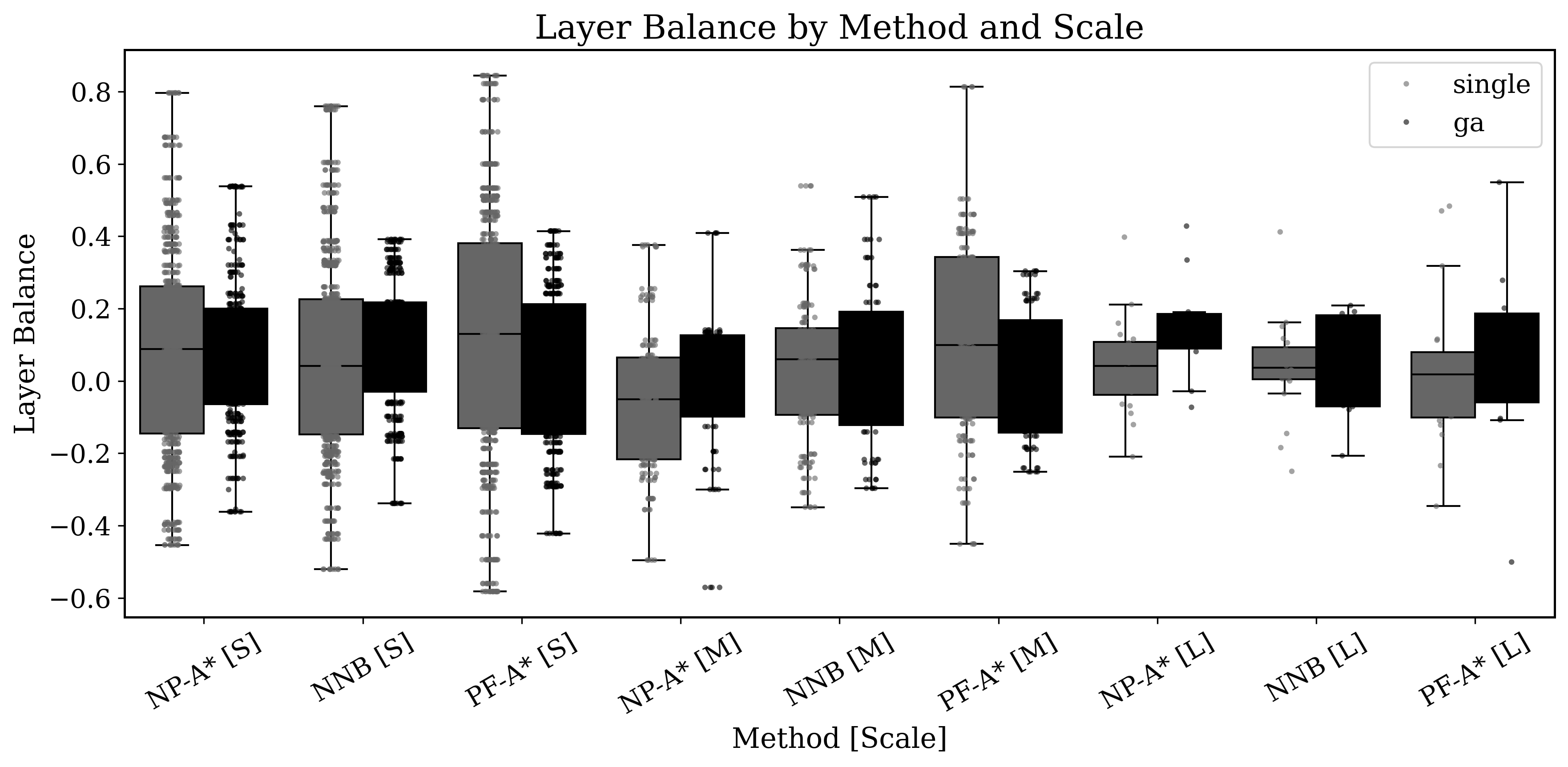}
  \caption{\textbf{Layer balance.} No contradicting imbalance; GA generally narrows dispersion.}
  \label{fig:layer-balance}
\end{figure}

\textbf{Switch Pattern Compliance (SPC).}
Fig.~\ref{fig:spc-box} visualizes SPC distributions. 
Aggregated means show that GA yields the cleanest patterns overall. Under GA, PF-A* attains the highest SPC at M, while NP-A* is marginally higher at S and L
(PF-A*: $\{0.737,\,\mathbf{0.838},\allowbreak\,0.791\}$ vs.\ NP-A*: $\{0.748,\,0.822,\,\mathbf{0.800}\}$; NNB: $\{0.715,\,0.709,\,0.444\}$).
In single mode, NP-A* is higher on S/M (NP-A*: $\{0.659,\,0.689,\,0.475\}$; PF-A*: $\{0.366,\,0.400,\,0.490\}$), while PF-A* catches up at L.
These trends are consistent with Section~\ref{sec:results-space}: PF-A* achieves near-perfect spacing MAE with competitive density, whereas SPC emphasizes minimum-gap compliance and shows NP-A* slightly ahead at S/L under GA.

\textbf{Switch Exit Freedom (SEF).}
Fig.~\ref{fig:sef-box} shows that NNB typically has the largest SEF 
(single: $\{0.596,\allowbreak\,0.538,\allowbreak\,0.463\}$; 
GA: $\{0.488,\allowbreak\,0.431,\allowbreak\,0.408\}$), 
followed by NP-A* (single: $\{0.345,\allowbreak\,0.296,\allowbreak\,0.241\}$; 
GA: $\{0.334,\allowbreak\,0.267,\allowbreak\,0.252\}$) and PF-A* 
(single: $\{0.133,\allowbreak\,0.131,\allowbreak\,0.179\}$; 
GA: $\{0.356,\allowbreak\,0.319,\allowbreak\,0.323\}$).

However, higher local exit openness (SEF) in NNB does not translate into either stronger controllability or robustness, reinforcing the need for structured switch placement.

\textbf{Layer balance.}
Fig.~\ref{fig:layer-balance} shows no anomalous imbalance that would contradict the main conclusions; GA generally tightens dispersion across methods.

\begin{figure}[t]
  \centering
  \includegraphics[width=\linewidth]{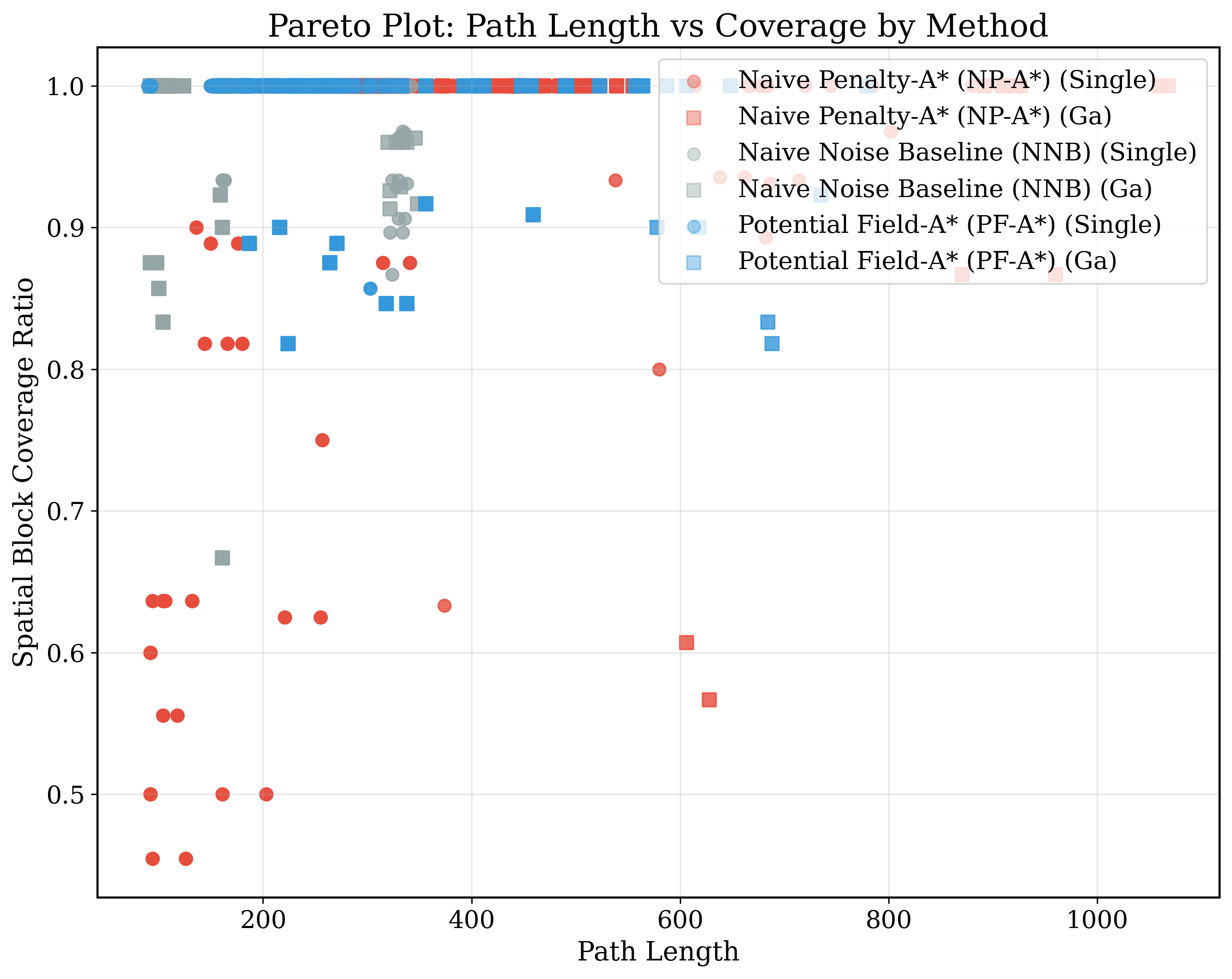}
  \caption{\textbf{Pareto: length vs.\ coverage.} PF-A* lies on a strong frontier; NP-A* reaches extreme coverage with longer paths; NNB is dominated.}
  \label{fig:pareto}
\end{figure}

\textbf{Length–coverage Pareto fronts.}
Fig.~\ref{fig:pareto} visualizes the trade-off between path length and spatial coverage across methods. PF-A* occupies a favorable frontier, achieving high coverage at shorter lengths; NP-A* can reach extreme coverage at the cost of longer paths; NNB is dominated. These trends are consistent with R4 in Section~\ref{sec:results-space}.

\section{Additional Quantitative Results for Time}
\label{app:time}

To complement the main results, we report extended statistical significance tests, efficiency plots, and raw performance tables. We conducted pairwise significance tests (two-sided Mann–Whitney U tests with Holm–Bonferroni correction, reporting Cliff's $\delta$) across methods at each scale (Small, Medium, Large) and mode (Single vs.\ GA). The results show that, across all tested settings, TEG-A* significantly outperforms the Static Backbone baseline in weighted scores, and that TEG-DP achieves the highest scores overall. In particular, TEG-DP significantly outperforms TEG-A* in all but two low-complexity settings (Small/Single and Medium/Single), where effect sizes are small and not statistically significant. Overall, this pattern supports our intended ordering:
\[
\text{Static Backbone} \;<\; \text{TEG-A*} \;<\; \text{TEG-DP}.
\]

Table~\ref{tab:significance} shows that pairwise differences are significant after Holm-Bonferroni correction in all cases except TEG-A* vs.\ TEG-DP under S-Single and M-Single (small, non-significant effects). GA further amplifies the gap, especially at L, while runtime exhibits the expected cost–quality trade-off (DP best quality at higher cost; A* a faster compromise; Static a lightweight ablation).

\begin{table}[t]
  \centering
  \small
  \caption{\textbf{Time direction: nonparametric significance and effect sizes.}
  Two-sided Mann–Whitney U with Holm–Bonferroni correction across key metrics; entries report Cliff's $\delta$ (magnitude) and adjusted $p$-values ($p_{\mathrm{adj}}$).}
  \label{tab:significance}
  \setlength{\tabcolsep}{5pt}
  \begin{tabular}{l l l c c}
    \toprule
    \textbf{Scale} & \textbf{Mode} & \textbf{Comparison} & \textbf{Cliff's $\delta$ (mag.)} & \textbf{$p_{\mathrm{adj}}$} \\
    \midrule
    S & Single & Static vs.\ TEG-A*  & $-0.997$ (\textit{large}) & $<\!0.001$ \\
    S & Single & TEG-A* vs.\ TEG-DP  & $0.236$ (\textit{small}) & $0.372$ \\
    S & GA     & Static vs.\ TEG-A*  & $-0.979$ (\textit{large}) & $<\!0.001$ \\
    S & GA     & TEG-A* vs.\ TEG-DP  & $-1.000$ (\textit{large}) & $<\!0.001$ \\
    \midrule
    M & Single & Static vs.\ TEG-A*  & $-0.992$ (\textit{large}) & $<\!0.001$ \\
    M & Single & TEG-A* vs.\ TEG-DP  & $-0.250$ (\textit{small}) & $0.854$ \\
    M & GA     & Static vs.\ TEG-A*  & $-0.867$ (\textit{large}) & $<\!0.001$ \\
    M & GA     & TEG-A* vs.\ TEG-DP  & $-1.000$ (\textit{large}) & $<\!0.001$ \\
    \midrule
    L & Single & Static vs.\ TEG-A*  & $-0.750$ (\textit{large}) & $0.002$ \\
    L & Single & TEG-A* vs.\ TEG-DP  & $-1.000$ (\textit{large}) & $<\!0.001$ \\
    L & GA     & Static vs.\ TEG-A*  & $-0.500$ (\textit{large}) & $<\!0.001$ \\
    L & GA     & TEG-A* vs.\ TEG-DP  & $-1.000$ (\textit{large}) & $<\!0.001$ \\
    \bottomrule
  \end{tabular}

  \vspace{2pt}
  \raggedright\footnotesize
  \textit{Notes.} Magnitude thresholds: negligible $|\delta|<0.11$, small $0.11{\le}|\delta|<0.28$, medium $0.28{\le}|\delta|<0.43$, large $|\delta|\ge 0.43$.
  Except for TEG-A* vs.\ TEG-DP at S-Single and M-Single (small, non-significant), all comparisons remain significant after Holm-Bonferroni correction.

\end{table}

\begin{table}[!t]
  \centering
  \caption{Time direction: runtime and weighted score across scale/method/mode (mean $\pm$ std).}
  \label{tab:time-score-combined}
  \begin{tabular}{c c c r@{$\,\pm\,$}l r@{$\,\pm\,$}l}
    \toprule
    Scale & Method & Mode & \multicolumn{2}{c}{Runtime (s)} & \multicolumn{2}{c}{Score} \\
    \midrule
    S & Static Backbone & Single & 0.19 & 0.17 & 8.22 & 2.65 \\
    S & Static Backbone & GA     & 34.39 & 5.75 & 13.56 & 1.68 \\
    S & TEG-A*          & Single & 0.17 & 0.22 & 9.19 & 6.09 \\
    S & TEG-A*          & GA     & 15.79 & 2.39 & 22.27 & 2.42 \\
    S & TEG-DP          & Single & 1.62 & 0.16 & 9.19 & 7.32 \\
    S & TEG-DP          & GA     & 159.23 & 14.77 & 26.10 & 9.31 \\
    \midrule
    M & Static Backbone & Single & 0.21 & 0.17 & 7.87 & 1.67 \\
    M & Static Backbone & GA     & 31.59 & 4.70 & 13.18 & 1.76 \\
    M & TEG-A*          & Single & 0.18 & 0.24 & 8.97 & 7.49 \\
    M & TEG-A*          & GA     & 15.76 & 2.33 & 22.22 & 2.84 \\
    M & TEG-DP          & Single & 2.86 & 0.34 & 11.84 & 14.68 \\
    M & TEG-DP          & GA     & 283.76 & 27.85 & 48.43 & 6.45 \\
    \midrule
    L & Static Backbone & Single & 0.31 & 0.20 & 7.81 & 1.77 \\
    L & Static Backbone & GA     & 30.20 & 3.82 & 13.20 & 2.33 \\
    L & TEG-A*          & Single & 0.25 & 0.18 & 13.30 & 3.51 \\
    L & TEG-A*          & GA     & 16.66 & 2.52 & 22.42 & 4.18 \\
    L & TEG-DP          & Single & 6.42 & 0.68 & 20.35 & 9.58 \\
    L & TEG-DP          & GA     & 576.72 & 35.00 & 84.04 & 12.99 \\
    \bottomrule
  \end{tabular}
\end{table}

\begin{figure*}[t]
  \centering
  \includegraphics[width=\textwidth]{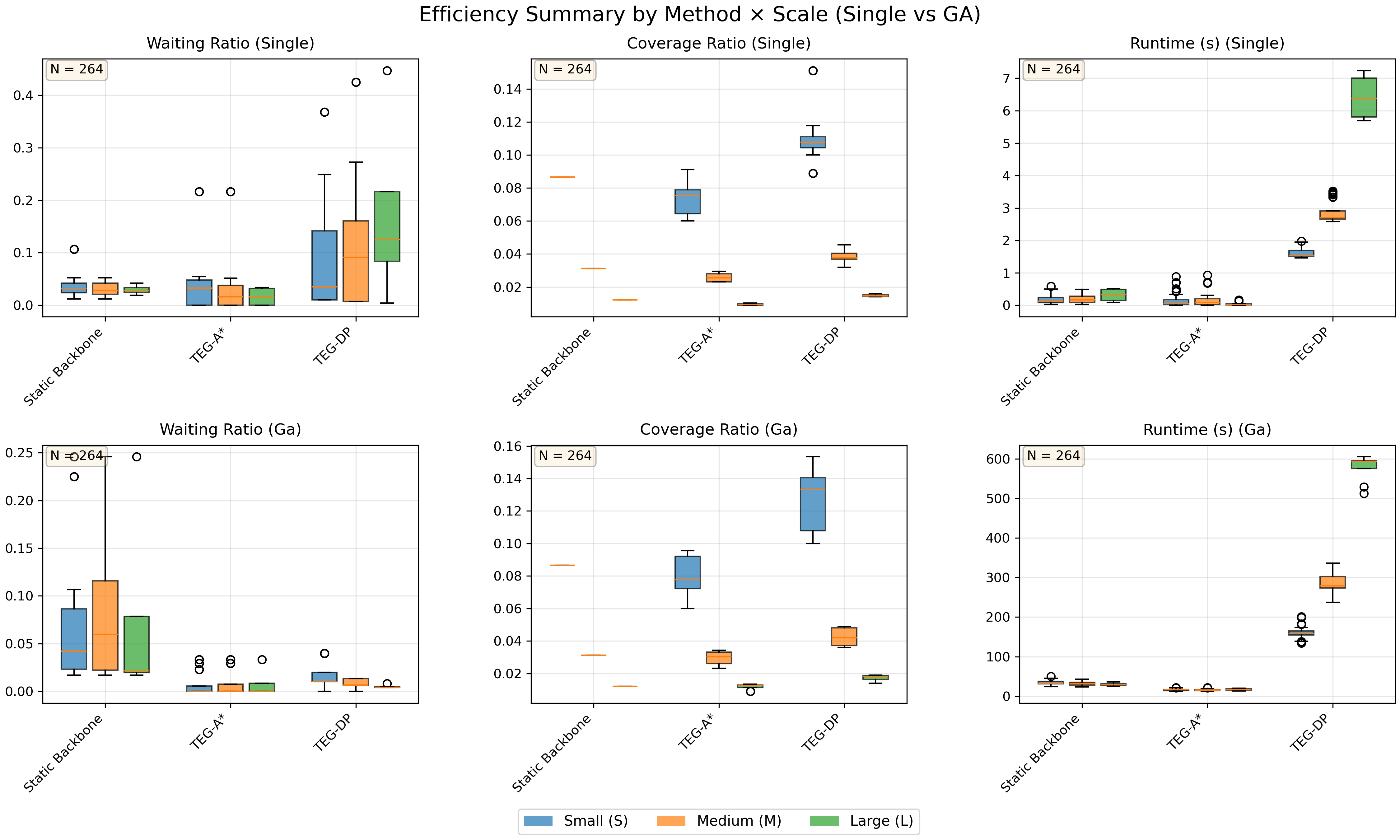}
  \caption{Efficiency summary across methods, scales, and modes.}
  \label{fig:efficiency-summary}
\end{figure*}

Figure~\ref{fig:efficiency-summary} provides the extended efficiency summary across all methods, scales, and modes. We observe consistent trends:
\begin{itemize}
  \item TEG-DP exhibits the highest ride ratio and coverage ratio, but with longer runtimes.
  \item TEG-A* achieves intermediate results, validating it as a middle ground between Static Backbone and DP.
  \item Static Backbone shows the lowest coverage and ride ratios, confirming its role as a minimal baseline.
\end{itemize}

Table~\ref{tab:time-score-combined} reports average runtime per method, scale, and mode. As expected, DP incurs the largest runtime cost due to its dynamic programming formulation, particularly under GA, but scales reliably. A* remains tractable across all scales in our experiments, with runtimes growing only moderately compared to DP. Table~\ref{tab:time-score-combined} also reports the aggregated weighted score means and standard deviations. Overall, the results support the ranking DP $>$ A* $>$ Static Backbone: both TEG variants consistently outperform the Static Backbone baseline across all scales and modes, and TEG-DP typically dominates TEG-A*, with only small, non-significant advantages over TEG-A* in the Small/Single and Medium/Single settings.

\end{document}